\documentclass{article}

\PassOptionsToPackage{numbers,sort&compress}{natbib}

\usepackage[final]{neurips_2019}

\usepackage[utf8]{inputenc} 
\usepackage[T1]{fontenc}    
\usepackage{hyperref}       
\usepackage{url}            
\usepackage{booktabs}       
\usepackage{amsfonts}       
\usepackage{nicefrac}       
\usepackage{microtype}      

\usepackage{color}
\definecolor{darkblue}{rgb}{0,0.0,0.55}
\hypersetup{ 
	colorlinks = true,
	citecolor  = darkblue,
	linkcolor  = darkblue,
	citecolor  = darkblue,
	filecolor  = darkblue,
	urlcolor   = darkblue,
      }

\usepackage{paralist}
\usepackage{cy}

\newcommand{\htanhact}{\sigma_{\rm{H}}}
\newcommand{\relulikeact}{\sigma_{\rm{R}}}
\newcommand{\gateact}{\sigma_{\rm{G}}}

\newcommand{\erisk}{\mathfrak{R}}
\newcommand{\btheta}{{\boldsymbol{\theta}}}
\newcommand{\bW}{{\boldsymbol{W}}}
\newcommand{\bb}{{\boldsymbol{b}}}
\newcommand{\bc}{{\boldsymbol{c}}}
\newcommand{\bV}{{\boldsymbol{V}}}
\newcommand{\bU}{{\boldsymbol{U}}}
\newcommand{\bxi}{{\boldsymbol{\xi}}}
\newcommand{\bDelta}{{\boldsymbol{\Delta}}}
\newcommand{\bdelta}{{\boldsymbol{\delta}}}
\newcommand{\bzeta}{{\boldsymbol{\zeta}}}
\newcommand{\vect}{\mathop{\rm vec}}

\newcommand{\Wmatent}[3]{\bW^{#1}_{{#2},{#3}}}
\newcommand{\Wmatrow}[2]{\bW^{#1}_{{#2},:}}
\newcommand{\bveccomp}[2]{\bb^{#1}_{#2}}
\newcommand{\unitvec}{{\boldsymbol{e}}}
\newcommand{\epoch}{E}

\title{Small ReLU networks are powerful memorizers: \\a tight analysis of memorization capacity}

\author{
	Chulhee Yun \\
	MIT\\
	Cambridge, MA 02139 \\
	\texttt{chulheey@mit.edu} \\
	\And
	Suvrit Sra \\
	MIT\\
	Cambridge, MA 02139 \\
	\texttt{suvrit@mit.edu}
	\And
	Ali Jadbabaie \\
	MIT\\
	Cambridge, MA 02139 \\
	\texttt{jadbabai@mit.edu}
}

\begin{document}

\maketitle

\begin{abstract}
We study finite sample expressivity, i.e., memorization power of ReLU networks. Recent results require $N$ hidden nodes to memorize/interpolate arbitrary $N$ data points. In contrast, by exploiting depth, we show that 3-layer ReLU networks with $\Omega(\sqrt{N})$ hidden nodes can perfectly memorize most datasets with $N$ points. We also prove that width $\Theta(\sqrt{N})$ is \emph{necessary and sufficient} for memorizing $N$ data points, proving tight bounds on memorization capacity. The sufficiency result can be extended to deeper networks; we show that an $L$-layer network with $W$ parameters in the hidden layers can memorize $N$ data points if $W = \Omega(N)$. Combined with a recent upper bound $O(WL\log W)$ on VC dimension, our construction is nearly tight for any fixed $L$. Subsequently, we analyze memorization capacity of residual networks under a general position assumption; we prove results that substantially reduce the known requirement of $N$ hidden nodes. Finally, we study the dynamics of stochastic gradient descent (SGD), and show that when initialized near a memorizing global minimum of the empirical risk, SGD quickly finds a nearby point with much smaller empirical risk.
\end{abstract}

\vspace*{-5pt}
\section{Introduction}
\vspace*{-5pt}
Recent results in deep learning indicate that over-parameterized neural networks can memorize arbitrary datasets~\citep{zhang2017understanding, arpit2017closer}. This phenomenon is closely related to the expressive power of neural networks, which have been long studied as universal approximators~\citep{cybenko1989approximation, hornik1989multilayer,funahashi1989approximate}. 
These results suggest that sufficiently large neural networks are expressive enough to fit any dataset perfectly. 

With the widespread use of deep networks, recent works have focused on better understanding the power of depth~\citep{delalleau2011shallow,telgarsky2015representation, telgarsky2016benefits,eldan2016power,safran2017depth,yarotsky2017error,yarotsky2018optimal,liang2017deep,lu2017expressive,rolnick2018power}.
However, most existing results consider expressing \emph{functions} (i.e., infinitely many points) rather than finite number of observations; thus, they do not provide a precise understanding the memorization ability of finitely large networks.

When studying finite sample memorization, several questions arise: Is a neural network capable of memorizing arbitrary datasets of a given size? How large must a neural network be to possess such capacity? These questions are the focus of this paper, and we answer them by studying \emph{universal finite sample expressivity} and \emph{memorization capacity}; these concepts are formally defined below.

\begin{definition} We define (universal) \textbf{finite sample expressivity} of a neural network $f_\btheta(\cdot)$ (parametrized by $\btheta$) as the network's ability to satisfy the following condition:
  \begin{quote}
    \vspace*{-5pt}
    \textbf{For all} inputs $\{x_i\}_{i=1}^N \in \reals^{d_x \times N}$ and \textbf{for all} $\{y_i\}_{i=1}^N \in [-1, +1]^{d_y \times N}$, \textbf{there exists} a parameter $\btheta$ such that $f_{\btheta} (x_i)=y_i$ for $1\le i \le N$.
    \vspace*{-5pt}
  \end{quote}
  We define \textbf{memorization capacity} of a network to be the maximum value of $N$ for which the network has finite sample expressivity when $d_y = 1$.
\end{definition}

Memorization capacity is related to, but is different from \textbf{VC dimension} of neural networks \citep{bartlett1999almost, bartlett2019nearly}. Recall the definition of VC dimension of a neural network $f_\btheta(\cdot)$:
\begin{quote}
  \vspace*{-5pt}
  The maximum value $N$ such that \textbf{there exists} a dataset $\{x_i\}_{i=1}^N \in \reals^{d_x \times N}$ such that \textbf{for all} $\{y_i\}_{i=1}^N \in \{\pm1\}^N$ \textbf{there exists} $\btheta$ such that $f_{\btheta} (x_i)=y_i$ for $1\le i\le N$.
  \vspace*{-8pt}
\end{quote}
Notice that the key difference between memorization capacity and VC dimension is in the quantifiers in front of the $x_i$'s. Memorization capacity is always less than or equal to VC dimension, which means that an upper bound on VC dimension is also an upper bound on memorization capacity.

The study of finite sample expressivity and memorization capacity of neural networks  has a long history, dating back to the days of perceptrons~\citep{cover1965geometrical, nilsson1965learning, baum1988capabilities, huang1991bounds, yamasaki1993lower, sontag1997shattering, kowalczyk1997estimates, huang1998upper, huang2003learning}; however, the older studies focus on shallow networks with traditional activations such as sigmoids, delivering limited insights for deep ReLU networks. Since the advent of deep learning, some recent results on modern architectures appeared, e.g., fully-connected neural networks (FNNs)~\citep{zhang2017understanding}, residual networks (ResNets)~\citep{hardt2017identity}, and convolutional neural networks (CNNs) \citep{nguyen2017losscnn}.
However, they impose assumptions on architectures that are neither practical nor realistic. For example, they require a hidden layer as wide as the number of data points $N$~\citep{zhang2017understanding, nguyen2017losscnn}, or as many hidden nodes as $N$~\citep{hardt2017identity}, causing their theoretical results to be applicable only to very large neural networks; this can be unrealistic especially when $N$ is large.

\vspace*{-5pt}
\subsection{Summary of our contributions}
\vspace*{-5pt}
Before stating our contributions, a brief comment on ``network size'' is in order. The size of a neural network can be somewhat vague; it could mean width/depth, the number of edges, or the number of hidden nodes. We use ``size'' to refer to the number of hidden nodes in a network. This also applies to notions related to size; e.g., by a ``small network'' we mean a network with a small number of hidden nodes. For other measures of size such as width, we will use the words explicitly.

%

\vspace*{-4pt}
\paragraph{1. Finite sample expressivity of neural networks.}
Our first set of results is on the finite sample expressivity of FNNs (Section~\ref{sec:fnns}), under the assumption of distinct data point $x_i$'s.
For simplicity, we only summarize our results for ReLU networks, but they include hard-tanh networks as well.
\begin{list}{{\tiny$\bullet$}}{\leftmargin=1.8em}
  \setlength{\itemsep}{2pt}
  \vspace*{-4pt}
\item Theorem~\ref{thm:htan3l} shows that any 3-layer (i.e., 2-hidden-layer) ReLU FNN with hidden layer widths $d_1$ and $d_2$ can fit \emph{any arbitrary} dataset if $d_1 d_2 \geq 4 N d_y$, where $N$ is the number of data points and $d_y$ is the output dimension. For scalar outputs, this means $d_1 = d_2 = 2 \sqrt N$ suffices to fit arbitrary data. This width requirement is significantly smaller than existing results on ReLU.
  
\item The improvement is more dramatic for classification. If we have $d_y$ classes, Proposition~\ref{prop:htanclsf4l} shows that a 4-layer ReLU FNN with hidden layer widths $d_1$, $d_2$, and $d_3$ can fit any dataset if $d_1 d_2 \geq 4N$ and $d_3 \geq 4 d_y$. This means that $10^6$ data points in $10^3$ classes (e.g., ImageNet) can be memorized by a 4-layer FNN with hidden layer widths 2k-2k-4k. 
  
\item For $d_y=1$, note that Theorem~\ref{thm:htan3l} shows a lower bound of $\Omega(d_1 d_2)$ on memorization capacity. We prove a matching upper bound in Theorem~\ref{thm:neg}: we show that for shallow neural networks (2 or 3 layers), lower bounds on memorization capacity are tight.
  
\item Proposition~\ref{thm:htanml} extends Theorem~\ref{thm:htan3l} to deeper and/or narrower networks, and shows that if the sum of the number of edges between pairs of adjacent layers satisfies $d_{l_1} d_{l_1 +1} + \cdots +  d_{l_m} d_{l_m +1} = \Omega(Nd_y)$, then universal finite sample expressivity holds. This gives a lower bound $\Omega(W)$ on memorization capacity, where $W$ is the number of edges in the network. Due to an upper bound $O(WL \log W)$ ($L$ is depth) on VC dimension \citep{bartlett2019nearly}, our lower bound is almost tight for fixed $L$.
\end{list}

Next, in Section~\ref{sec:genpos}, we focus on classification using ResNets; here $d_x$ denotes the input dimension and $d_y$ the number of classes. We assume here that data lies in general position. 
\begin{list}{\tiny{$\bullet$}}{\leftmargin=1.8em}
\vspace*{-5pt}
\item Theorem~\ref{thm:genposresnet} proves that deep ResNets with $\frac{4N}{d_x}+6d_y$ ReLU hidden nodes can memorize arbitrary datasets. 
Using the same proof technique, we also show in Corollary~\ref{cor:genposfnn} that a 2-layer ReLU FNN can memorize arbitrary classification datasets if $d_1 \geq \frac{4N}{d_x}+4d_y$. With the general position assumption, we can reduce the existing requirements of $N$ to a more realistic number.
\end{list}

  \vspace*{-3pt}
\paragraph{2. Trajectory of SGD near memorizing global minima.}
Finally, in Section~\ref{sec:gd} we study the behavior of stochastic gradient descent (SGD) on the empirical risk of universally expressive FNNs.
\begin{list}{\tiny{$\bullet$}}{\leftmargin=1.8em}
\setlength{\itemsep}{0pt}
\vspace*{-5pt}
\item Theorem~\ref{thm:memorizer} shows that for \emph{any} differentiable global minimum that memorizes, SGD initialized close enough (say $\epsilon$ away) to the minimum, quickly finds a point that has empirical risk $O(\epsilon^4)$ and is at most $2 \epsilon$ far from the minimum. We emphasize that this theorem holds not only for memorizers explicitly constructed in Sections~\ref{sec:fnns} and  \ref{sec:genpos}, but for \textit{all} global minima that memorize. We note that we analyze without replacement SGD that is closer to practice than the simpler with-replacement version~\citep{haochen2018random,shamir2016without}; thus, our analysis may be of independent interest in optimization. 
  \vspace*{-3pt}
\end{list}
\vspace{-5pt}
\subsection{Related work}
\vspace*{-5pt}
\paragraph{Universal finite sample expressivity of neural networks.}
Literature on finite sample expressivity and memorization capacity of neural networks dates back to the 1960s. Earlier results~\citep{cover1965geometrical, nilsson1965learning, baum1988capabilities, sontag1997shattering, kowalczyk1997estimates} study memorization capacity of linear threshold networks.

Later, results on 2-layer FNNs with sigmoids~\citep{huang1991bounds} and other bounded activations~\citep{huang1998upper} show that $N$ hidden nodes are sufficient to memorize $N$ data points. It was later shown that the requirement of $N$ hidden nodes can be improved by exploiting depth~\citep{yamasaki1993lower, huang2003learning}. Since these two works are highly relevant to our own results, we defer a detailed discussion/comparison until we present the precise theorems (see Sections~\ref{sec:maindis} and \ref{sec:deeper}).

With the advent of deep learning, there have been new results on modern activation functions and architectures. \citet{zhang2017understanding} prove that one-hidden-layer ReLU FNNs with $N$ hidden nodes can memorize $N$ real-valued data points. \citet{hardt2017identity} show that deep ResNets with $N + d_y$ hidden nodes can memorize arbitrary $d_y$-class classification datasets. \citet{nguyen2017losscnn} show that deep CNNs with one of the hidden layers as wide as $N$ can memorize $N$ real-valued data points.

\citet{soudry2016no} show that under a dropout noise setting, the training error is zero at every differentiable local minimum, for almost every dataset and dropout-like noise realization.
However, this result is not comparable to ours because they assume that there is a multiplicative ``dropout noise'' at each hidden node and each data point. At $i$-th node of $l$-th layer, the slope of the activation function for the $j$-th data point is either $\epsilon_{i,l}^{(j)} \cdot 1$ (if input is positive) or $\epsilon_{i,l}^{(j)} \cdot s$ (if input is negative, $s \neq 0$), where $\epsilon_{i,l}^{(j)}$ is the multiplicative random (e.g., Gaussian) dropout noise. Their theorem statements hold for all realizations of these dropout noise factors \emph{except a set of measure zero}. In contrast, our setting is free of these noise terms, and hence corresponds to a \emph{specific} realization of such $\epsilon_{i,l}^{(n)}$'s.


\vspace*{-5pt}
\paragraph{Convergence to global minima.}
There exist numerous papers that study convergence of gradient descent or SGD to global optima of neural networks. Many previous results \citep{tian2017analytical, brutzkus2017globally, zhong2017recovery, soltanolkotabi2017learning, li2017convergence, du2017gradient, zhang2018learning} study settings where data points are sampled from a distribution (e.g., Gaussian), and labels are generated from a ``teacher network'' that has the same architecture as the one being trained (i.e., realizability). Here, the goal of training is to recover the unknown (but fixed) true parameters. In comparison, we consider arbitrary datasets and networks, under a mild assumption (especially for overparametrized networks) that the network can memorize the data; the results are not directly comparable.
Others~\citep{brutzkus2018sgd, wang2018learning} study SGD on hinge loss under a bit strong assumption that the data is linearly separable.

Other recent results \citep{li2018learning, du2018gradientB, du2018gradientA, allen2018convergence, zou2018stochastic} focus on over-parameterized neural networks. In these papers, the widths of hidden layers are assumed to be huge, of polynomial order in $N$, such as $\Omega(N^4)$, $\Omega(N^6)$ or even greater. Although these works provide insights on how GD/SGD finds global minima easily, their width requirement is still far from being realistic.


A recent work~\citep{zhou2019sgd} provides a mixture of observation and theory about convergence to global minima. The authors assume that networks can memorize the data, and that SGD follows a star-convex path to global minima, which they validate through experiments. Under these assumptions, they prove convergence of SGD to global minimizers. We believe our result is complementary: we provide sufficient conditions for networks to memorize the data, and our result does not assume anything about SGD's path but proves that SGD can find a point close to the global minimum.


\vspace{-5pt}
\paragraph{Remarks on generalization.}
The ability of neural networks to memorize and generalize at the same time has been one of the biggest mysteries of deep learning \citep{zhang2017understanding}. Recent results on interpolation and ``double descent'' phenomenon indicate that memorization may not necessarily mean lack of generalization \citep{liang2018just, belkin2018does, belkin2018reconciling, bartlett2019benign, mei2019generalization, liang2019risk}.
We note that our paper focuses mainly on the ability of neural networks to memorize the training dataset, and that our results are separate from the discussion of generalization.

\vspace*{-8pt}
\section{Problem setting and notation}
\vspace*{-6pt}
\label{sec:prelim}
In this section, we introduce the notation used throughout the paper.
For integers $a$ and $b$, $a < b$, we denote $[a] \defeq \{1,\dots,a\}$ and $[a:b] \defeq \{a, a+1, \dots, b \}$. We denote $\{(x_i, y_i)\}_{i=1}^N$ the set of training data points, and our goal is to choose the network parameters $\btheta$ so that the network output $f_{\btheta} (x_i)$ is equal to $y_i$, for all $i \in [n]$. Let $d_x$ and $d_y$ denote input and output dimensions, respectively. Given input $x \in \reals^{d_x}$, an $L$-layer fully-connected neural network computes output $f_{\btheta}(x)$ as follows:
\begin{align*}
a^{0}(x) &= x,\quad\\
z^{l}(x) &= \bW^{l} a^{l-1}(x) + \bb^{l},\quad
a^{l}(x) = \sigma(z^{l}(x)),\quad\text{ for } l \in [L-1],\\
f_{\btheta}(x) &= \bW^{L} a^{L-1}(x) + \bb^{L}.
\end{align*}
Let $d_l$ (for $l \in [L-1]$) denote the width of $l$-th hidden layer. For convenience, we write $d_0 \defeq d_x$ and $d_L \defeq d_y$. Here, $z^{l} \in \reals^{d_l}$ and $a^{l} \in \reals^{d_l}$ denote the input and output ($a$ for activation) of the $l$-th hidden layer, respectively. The output of a hidden layer is the entry-wise map of the input by the activation function $\sigma$. The bold-cased symbols denote parameters: $\bW^{l} \in \reals^{d_l \times d_{l-1}}$ is the weight matrix, and $\bb^{l} \in \reals^{d_l}$ is the bias vector. 
We define $\btheta \defeq (\bW^{l}, \bb^{l})_{l=1}^L$ to be the collection of all parameters. We write the network output as $f_{\btheta} (\cdot)$ to emphasize that it depends on parameters $\btheta$.

Our results in this paper consider piecewise linear activation functions. Among them, Sections~\ref{sec:fnns} and \ref{sec:genpos} consider ReLU-like ($\relulikeact$) and hard-tanh ($\htanhact$) activations, defined as follows:
\begin{equation*}
\relulikeact(t) \defeq \begin{cases}
s_+ t & t \geq 0,\\
s_- t & t < 0,
\end{cases}\ \ 
\htanhact(t) \defeq \begin{cases}
-1 & t \leq -1,\\
t & t \in (-1,1],\\
1 & t > 1,
\end{cases}
= \frac{\relulikeact(t+1) - \relulikeact(t-1)-s_+-s_-}{s_+-s_-},
\end{equation*}
where $s_+ > s_- \geq 0$. Note that $\relulikeact$ includes ReLU and Leaky ReLU.
Hard-tanh activation ($\htanhact$) is a piecewise linear approximation of tanh. 
Since $\htanhact$ can be represented with two $\relulikeact$, any results on hard-tanh networks can be extended to ReLU-like networks with twice the width.


\vspace*{-5pt}
\section{Finite sample expressivity of FNNs}
\label{sec:fnns}
\vspace*{-5pt}
In this section, we study universal finite sample expressivity of FNNs.
For the training dataset, we make the following mild assumption that ensures consistent labels:
\begin{assumption}
\label{asm:data}
In the dataset $\{(x_i, y_i)\}_{i=1}^N$ assume that all $x_i$'s are distinct and all $y_i \in [-1,1]^{d_y}$. 
\end{assumption}

\vspace*{-8pt}
\subsection{Main results}
\vspace*{-5pt}
We first state the main theorems on shallow FNNs showing tight lower and upper bounds on memorization capacity. Detailed discussion will follow in the next subsection.
\begin{theorem}
	\label{thm:htan3l}
	Consider any dataset $\{(x_i, y_i)\}_{i=1}^N$ that satisfies Assumption~\ref{asm:data}. 
	If
	\begin{itemize}
		\vspace*{-5pt}
		\setlength{\itemsep}{0pt}
		\item a 3-layer hard-tanh FNN $f_\btheta$ satisfies 
		$4\lfloor d_1/2 \rfloor \lfloor d_2/(2 d_y) \rfloor \geq N$; or
		\item a 3-layer ReLU-like FNN $f_\btheta$ satisfies 
		$4\lfloor d_1/4 \rfloor \lfloor d_2/(4d_y) \rfloor \geq N$,
		\vspace*{-5pt}
	\end{itemize}
	then there exists a parameter $\btheta$ such that $y_i = f_{\btheta} (x_i)$ for all $i \in [N]$.
\end{theorem}
Theorem~\ref{thm:htan3l} shows that if $d_1 d_2 = \Omega(N d_y)$ then we can memorize arbitrary datasets; this means that $\Omega(\sqrt {N d_y})$ hidden nodes are sufficient for memorization, in contrary to $\Omega(N d_y)$ requirements of recent results.
By adding one more hidden layer, the next theorem shows that we can perfectly memorize any \textit{classification} dataset using $\Omega(\sqrt N + d_y)$ hidden nodes.
\begin{proposition}
	\label{prop:htanclsf4l}
	Consider any dataset $\{(x_i, y_i)\}_{i=1}^N$ that satisfies Assumption~\ref{asm:data}. Assume that $y_i \in \{0,1\}^{d_y}$ is the one-hot encoding of $d_y$ classes.
	Suppose one of the following holds:
	\begin{itemize}
		\vspace*{-5pt}
		\setlength{\itemsep}{0pt}
		\item a 4-layer hard-tanh FNN $f_\btheta$ satisfies 
		$4\lfloor d_1/2 \rfloor \lfloor d_2/2 \rfloor \geq N$, and $d_3 \geq 2 d_y$; or
		\item a 4-layer ReLU-like FNN $f_\btheta$ satisfies 
		$4\lfloor d_1/4 \rfloor \lfloor d_2/4 \rfloor \geq N$, and $d_3 \geq 4 d_y$.
		\vspace*{-5pt}
	\end{itemize}
	Then, there exists a parameter $\btheta$ such that $y_i = f_{\btheta} (x_i)$ for all $i \in [N]$.
\end{proposition}
Notice that for scalar regression ($d_y = 1$), Theorem~\ref{thm:htan3l} proves a lower bound on memorization capacity of 3-layer neural networks: $\Omega(d_1 d_2)$. The next theorem shows that this bound is in fact \emph{tight}.
\begin{theorem}
	\label{thm:neg}
	Consider FNNs with $d_y = 1$ and piecewise linear activation $\sigma$ with $p$ pieces.
	If 
	\begin{itemize}
		\vspace*{-5pt}
		\setlength{\itemsep}{0pt}
		\item a 2-layer FNN $f_\btheta$ satisfies $(p-1) d_1 + 2 < N$; or
		\item a 3-layer FNN $f_\btheta$ satisfies $p(p-1) d_1 d_2 + (p-1) d_2 + 2 < N$,
		\vspace*{-5pt}
	\end{itemize}
	then there exists a dataset $\{(x_i, y_i)\}_{i=1}^N$ satisfying Assumption~\ref{asm:data} such that for all $\btheta$, there exists $i \in [N]$ such that $y_i \neq f_{\btheta} (x_i)$.
\end{theorem}
Theorems~\ref{thm:htan3l} and \ref{thm:neg} together show \textbf{tight} lower and upper bounds $\Theta(d_1 d_2)$ on memorization capacity of 3-layer FNNs, which differ only in constant factors. Theorem~\ref{thm:neg} and the existing result on 2-layer FNNs \citep[Theorem~1]{zhang2017understanding} also show that the memorization capacity of 2-layer FNNs is $\Theta(d_1)$.

\paragraph{Proof ideas.}
The proof of Theorem~\ref{thm:htan3l} is based on an intricate construction of parameters. Roughly speaking, we construct parameters that make each data point have its unique activation pattern in the hidden layers; more details are in Appendix~\ref{sec:thmpf-htan3l}. 
The proof of Proposition~\ref{prop:htanclsf4l} is largely based on Theorem~\ref{thm:htan3l}. By assigning each class $j$ a unique real number $\rho_j$ (which is similar to the trick in \citet{hardt2017identity}), we modify the dataset into a 1-D regression dataset; we then fit this dataset using the techniques in Theorem~\ref{thm:htan3l}, and use the extra layer to recover the one-hot representation of the original $y_i$. Please see Appendix~\ref{sec:thmpf-htanclsf4l} for the full proof.
The main proof idea of Theorem~\ref{thm:neg} is based on counting the number of ``pieces'' in the network output $f_\btheta(x)$ (as a function of $x$), inspired by \citet{telgarsky2015representation}. For the proof, please see Appendix~\ref{sec:thmpf-neg}.


\vspace*{-5pt}
\subsection{Discussion}
\label{sec:maindis}

\vspace*{-5pt}
\paragraph{Depth-width tradeoffs for finite samples.}

Theorem~\ref{thm:htan3l} shows that if the two ReLU hidden layers satisfy $d_1 = d_2 = 2\sqrt{N d_y}$, then the network can fit a given dataset perfectly. Proposition~\ref{prop:htanclsf4l} is an improvement for classification, which shows that a 4-layer ReLU FNN can memorize any $d_y$-class classification data if $d_1  = d_2 = 2\sqrt N$ and $d_3 = 4d_y$.

As in other expressivity results, our results show that there are depth-width tradeoffs in the finite sample setting. For ReLU FNNs it is known that one hidden layer with $N$ nodes can memorize any scalar regression ($d_y=1$) dataset with $N$ points \citep{zhang2017understanding}. By adding a hidden layer, the hidden node requirement is reduced to $4\sqrt N$, and Theorem~\ref{thm:neg} also shows that $\Theta(\sqrt N)$ hidden nodes are \emph{necessary and sufficient}. Ability to memorize $N$ data points with $N$ nodes is perhaps not surprising, because weights of each hidden node can be tuned to memorize a single data point. In contrast, the fact that width-$2 \sqrt N$ networks can memorize is far from obvious; each hidden node must handle $\sqrt N/2$ data points on average, thus a more elaborate construction is required. 

For $d_y$-class classification, by adding one more hidden layer, the requirement is improved from $4\sqrt{N d_y}$ to $4\sqrt{N} + 4 d_y$ nodes. This again highlights the power of depth in expressive power.
Proposition~\ref{prop:htanclsf4l} tells us that we can fit ImageNet\footnote{after omitting the inconsistently labeled items} ($N \approx 10^6, d_y = 10^3$) with three ReLU hidden layers, using only 2k-2k-4k nodes.
This ``sufficient'' size for memorization is surprisingly smaller (disregarding optimization aspects) than practical networks.

\vspace*{-9pt}
\paragraph{Implications for ERM.}
It is widely observed in experiments that deep neural networks can achieve zero empirical risk, but a concrete understanding of this phenomenon is still elusive.
It is known that all local minima are global minima for empirical risk of linear neural networks~\citep{kawaguchi2016deep,yun2018global, zhou2018critical, laurent2018deep, yun2019small}, but this property fails to extend to nonlinear neural networks \citep{safran2017spurious, yun2019small}. This suggests that studying the gap between local minima and global minima could provide explanations for the success of deep neural networks. In order to study the gap, however, we have to know the risk value attained by global minima, which is already non-trivial even for shallow neural networks. 
In this regard, our theorems provide theoretical guarantees that even a shallow and narrow network can have zero empirical risk at global minima, \emph{regardless of data and loss functions}---e.g., in a regression setting, for a 3-layer ReLU FNN with $d_1 = d_2 = 2 \sqrt{N d_y}$ there exists a global minimum that has zero empirical risk.

\vspace*{-7pt}
\paragraph{The number of edges.}
We note that our results do not contradict the common ``insight'' that at least $N$ edges are required to memorize $N$ data points. Our ``small'' network means a small number of hidden nodes, and it still has more than $N$ edges.
The existing result \citep{zhang2017understanding} requires $(d_x+2)N$ edges, while our construction for ReLU requires $4N+(2d_x+6)\sqrt{N}+1$ edges, which is much fewer.


\vspace*{-7pt}
\paragraph{Relevant work on sigmoid.}
\citet{huang2003learning} proves that a 2-hidden-layer sigmoid FNNs with $d_1 = N/K+2K$ and $d_2 = K$, where $K$ is a positive integer, can approximate $N$ arbitrary distinct data points. 
The author first partitions $N$ data points into $K$ groups of size $N/K$ each. Then, from the fact that the sigmoid function is strictly increasing and non-polynomial, it is shown that if the weights between input and first hidden layer is sampled \emph{randomly}, then the output matrix of first hidden layer for each group is full rank with probability one. This is \emph{not} the case for ReLU or hard-tanh, because they have ``flat'' regions in which rank could be lost.
In addition, \citet{huang2003learning} requires extra $2K$ hidden nodes in $d_1$ that serve as ``filters'' which let only certain groups of data points pass through.
Our construction is not an extension of this result because we take a different strategy (Appendix~\ref{sec:thmpf-htan3l}); we carefully choose parameters (instead of sampling) that achieve memorization with $d_1 = N/K$ and $d_2 = K$ (in hard-tanh case) without the need of extra $2K$ nodes, which enjoys a smaller width requirement and allows for more flexibility in the architecture. Moreover, we provide a converse result (Theorem~\ref{thm:neg}) showing that our construction is rate-optimal in the number of hidden nodes.

%

\vspace*{-5pt}
\subsection{Extension to deeper and/or narrower networks}
\label{sec:deeper}
\vspace*{-5pt}

What if the network is deeper than three layers and/or narrower than $\sqrt N$? 
Our next theorem shows that universal finite sample expressivity is not limited to 3-layer neural networks, and still achievable by exploiting depth even for narrower networks.
\begin{proposition}
	\label{thm:htanml}
	Consider any dataset $\{(x_i, y_i)\}_{i=1}^N$ that satisfies Assumption~\ref{asm:data}. 
	For an $L$-layer FNN with hard-tanh activation \textup{($\htanhact$)}, assume that there exist indices $l_1, \dots, l_m \in [L-2]$ that satisfy
	\begin{itemize}
	\vspace*{-5pt}
	\setlength{\itemsep}{0pt}
	\item $l_j + 1 < l_{j+1}$ for $j \in [m-1]$,
	\item $4 \sum_{j=1}^{m} \left \lfloor \frac{d_{l_j}-r_j}{2} \right \rfloor 
	\left \lfloor \frac{d_{l_j+1}-r_j}{2d_y} \right \rfloor \geq N$, where $r_j = d_y \indic{j>1}+\indic{j<m}$, for $j \in [m]$,
	\item $d_k \geq d_y + 1$ for all $k \in \bigcup_{j \in [m-1]}[l_j + 2:l_{j+1}-1]$.
	\item $d_k \geq d_y$ for all $k \in [l_m + 2: L-1]$,
	\vspace*{-5pt}
	\end{itemize}
	where $\indic{\cdot}$ is 0-1 indicator function. Then, there exists $\btheta$ such that $y_i = f_{\btheta} (x_i)$ for all $i \in [N]$.
\end{proposition}
As a special case, note that for $L=3$ (hence $m=1$), the conditions boil down to that of Theorem~\ref{thm:htan3l}.
An immediate corollary of this fact is that the same result holds for ReLU(-like) networks with twice the width. 
Moreover, using the same proof technique as Proposition~\ref{prop:htanclsf4l}, this theorem can also be improved for classification datasets, by inserting one additional hidden layer between layer~$l_m+1$ and the output layer. Due to space limits, we defer the statement of these corollaries to Appendix~\ref{sec:thm-deferred}.

The proof of Proposition~\ref{thm:htanml} is in Appendix~\ref{sec:thmpf-htanml}. We use Theorem~\ref{thm:htan3l} as a building block and construct a network (see Figure~\ref{fig:fig2} in appendix) that fits a subset of dataset at each pair of hidden layers $l_j$--($l_j+1$).

If any two adjacent hidden layers satisfy $d_{l} d_{l+1} = \Omega(Nd_y)$, this network can fit $N$ data points ($m=1$), even when all the other hidden layers have only one hidden node. Even with networks narrower than $\sqrt {N d_y}$ (thus $m>1$), we can still achieve universal finite sample expressivity as long as there are $\Omega(N d_y)$ edges between disjoint pairs of adjacent layers. However, we have the ``cost'' $r_j$ in the width of hidden layers; this is because we fit subsets of the dataset using multiple pairs of layers. To do this, we need $r_j$ extra nodes to propagate input and output information to the subsequent layers. For more details, please refer to the proof.


Proposition~\ref{thm:htanml} gives a lower bound $\Omega(\sum_{l=1}^{L-2}d_l d_{l+1})$ on memorization capacity for $L$-layer networks. For fixed input/output dimensions, this is indeed $\Omega(W)$, where $W$ is the number of edges in the network. On the other hand, \citet{bartlett2019nearly} showed an upper bound $O(WL \log W)$ on VC dimension, which is also an upper bound on memorization capacity. Thus, for any fixed $L$, our lower bound is nearly tight. We conjecture that, as we have proved in 2- and 3-layer cases, the memorization capacity is $\Theta(W)$, independent of $L$; we leave closing this gap for future work. 

For sigmoid FNNs, \citet{yamasaki1993lower} claimed that a scalar regression dataset can be memorized if $d_x \lceil \frac{d_1}{2} \rceil + \lfloor \frac{d_1}{2} \rfloor \lceil \frac{d_2}{2}-1 \rceil + \cdots + \lfloor \frac{d_{L-2}}{2} \rfloor \lceil \frac{d_{L-1}}{2}-1 \rceil \geq N$.
However, this claim was made under the stronger assumption of data lying in general position (see Assumption~\ref{asm:datageneral}). Unfortunately, \citet{yamasaki1993lower} does not provide a full proof of their claim, making it impossible to validate veracity of their construction (and we could not find their extended manuscript elsewhere).

\vspace*{-5pt}
\section{Classification under the general position assumption}
\label{sec:genpos}
\vspace*{-5pt}
This section presents some results specialized in multi-class classification task under a slightly stronger assumption, namely the general position assumption. Since we are only considering classification in this section, we also assume that $y_i \in \{0,1\}^{d_y}$ is the one-hot encoding of $d_y$ classes. 
\begin{assumption}
\label{asm:datageneral}
For a finite dataset $\{(x_i, y_i)\}_{i=1}^N$, assume that no $d_x+1$ data points lie on the same affine hyperplane. In other words, the data point $x_i$'s are in general position.
\end{assumption}

We consider \textbf{residual networks} (ResNets), defined by the following architecture:
\begin{align*}
h^{0}(x) &= x,\\
h^{l}(x) &= h^{l-1}(x) + \bV^l \sigma (\bU^l h^{l-1}(x) + \bb^{l})+\bc^{l},~ l \in [L-1],\\
g_{\btheta}(x) &= \bV^L \sigma (\bU^L h^{L-1}(x) + \bb^{L})+\bc^{L},
\end{align*}
which is similar to the previous work by \citet{hardt2017identity}, except for extra bias parameters $\bc^{l}$. In this model, we denote the number hidden nodes in the $l$-th residual layer as $d_l$; e.g., $\bU^l \in \reals^{d_l \times d_x}$.

We now present a theorem showing that any dataset can be memorized with small ResNets.
\begin{theorem}
\label{thm:genposresnet}
Consider any dataset $\{(x_i, y_i)\}_{i=1}^N$ that satisfies Assumption~\ref{asm:datageneral}. Assume also that $d_x \geq d_y$.
Suppose one of the following holds:
\begin{itemize}
	\vspace*{-5pt}
	\setlength{\itemsep}{0pt}
	\item a hard-tanh ResNet $g_\btheta$ satisfies 
	$\sum_{l=1}^{L-1} d_l \geq \frac{2N}{d_x}+2d_y$ and $d_L \geq d_y$; or
	\item a ReLU-like ResNet $g_\btheta$ satisfies 
	$\sum_{l=1}^{L-1} d_l \geq \frac{4N}{d_x}+4d_y$ and $d_L \geq 2d_y$.
	\vspace*{-5pt}
\end{itemize}
Then, there exists $\btheta$ such that $y_i = g_{\btheta} (x_i)$ for all $i \in [N]$.
\end{theorem}
The previous work by
\citet{hardt2017identity} proves universal finite sample expressivity using $N+d_y$ hidden nodes (i.e., $\sum_{l=1}^{L-1} d_l \geq N$ and $d_L \geq d_y$) for ReLU activation, under the assumption that $x_i$'s are distinct \textit{unit} vectors. Note that neither this assumption nor Assumption~\ref{asm:datageneral} implies the other; however, our assumption is quite mild in the sense that for any given dataset, adding small random Gaussian noise to $x_i$'s makes the dataset satisfy the assumption, with probability 1.

The main idea for the proof is that under the general position assumption, for any choice of $d_x$ points there exists an affine hyperplane that contains \emph{only} these $d_x$ points. Each hidden node can choose $d_x$ data points and ``push'' them to the right direction, making perfect classification possible. We defer the details to Appendix~\ref{sec:thmpf-genpos}.
Using the same technique, we can also prove an improved result for 2-layer (1-hidden-layer) FNNs. The proof of the following corollary can be found in Appendix~\ref{sec:corpf-genposfnn}.
\begin{corollary}
\label{cor:genposfnn}
Consider any dataset $\{(x_i, y_i)\}_{i=1}^N$ that satisfies Assumption~\ref{asm:datageneral}. Suppose one of the following holds:
\begin{itemize}
	\vspace*{-5pt}
	\setlength{\itemsep}{0pt}
	\item a 2-layer hard-tanh FNN $f_\btheta$ satisfies 
	$d_1 \geq \frac{2N}{d_x}+2d_y$; or
	\item a 2-layer ReLU-like FNN $f_\btheta$ satisfies 
	$d_1 \geq \frac{4N}{d_x}+4d_y$.
\end{itemize}
Then, there exists $\btheta$ such that $y_i = f_{\btheta} (x_i)$ for all $i \in [N]$.
\end{corollary}

Our results show that under the general position assumption, perfect memorization is possible with only $\Omega(N/d_x+d_y)$ hidden nodes rather than $N$, in both ResNets and 2-layer FNNs. Considering that $d_x$ is typically in the order of hundreds or thousands, our results reduce the hidden node requirements down to more realistic network sizes. For example, consider CIFAR-10 dataset: $N = 50,000$, $d_x = 3,072$, and $d_y = 10$. Previous results require at least 50k ReLUs to memorize this dataset, while our results require 126 ReLUs for ResNets and 106 ReLUs for 2-layer FNNs.



\vspace*{-5pt}
\section{Trajectory of SGD near memorizing global minima}
\label{sec:gd}
\vspace*{-5pt}
In this section, we study the behavior of without-replacement SGD near memorizing global minima.

We restrict $d_y = 1$ for simplicity. We use the same notation as defined in Section~\ref{sec:prelim}, and introduce here some additional definitions. 
We assume that each activation function $\sigma$ is piecewise linear with at least two pieces (e.g., ReLU or hard-tanh). Throughout this section, we slightly abuse the notation $\btheta$ to denote the concatenation of vectorizations of all the parameters $(\bW^l, \bb^l)_{l=1}^L$.

We are interested in minimizing the empirical risk $\erisk(\btheta)$, defined as the following:
\begin{equation*}
\erisk(\btheta) \defeq \tfrac{1}{N} \sum\nolimits_{i=1}^N \ell (f_\btheta(x_i); y_i),
\end{equation*}
where $\ell(z; y) : \reals \mapsto \reals$ is the loss function parametrized by $y$. We assume the following:
\begin{assumption}
\label{asm:loss}
The loss function $\ell(z; y)$ is a strictly convex and three times differentiable function of $z$. Also, for any $y$, there exists $z \in \reals$ such that $z$ is a global minimum of $\ell(z;y)$.
\end{assumption}
Assumption~\ref{asm:loss} on $\ell$ is satisfied by standard losses such as squared error loss. Note that logistic loss does not satisfy Assumption~\ref{asm:loss} because the global minimum is not attained by any finite $z$.

Given the assumption on $\ell$, we now formally define the \textbf{memorizing} global minimum. 
\begin{definition}
A point $\btheta^*$ is a memorizing global minimum of $\erisk(\cdot)$ if $\ell'(f_{\btheta^*}(x_i);y_i) = 0$, $\forall i \in [N]$.
\end{definition}
\vspace*{-5pt}
By convexity, $\ell'(f_{\btheta^*}(x_i);y_i) = 0$ for all $i$ implies that $\erisk(\btheta)$ is (globally) minimized at $\btheta^*$. 
Also, existence of a memorizing global minimum of $\erisk$ implies that all global minima are memorizing.

Although $\ell$ is a differentiable function of $z$, the empirical risk $\erisk(\btheta)$ is not necessarily differentiable in $\btheta$ because we are using  piecewise linear activations. 
In this paper, we only consider differentiable points of $\erisk(\cdot)$; since nondifferentiable points lie in a set of measure zero and SGD never reaches such points in reality, this is a reasonable assumption.

We consider minimizing the empirical risk $\erisk(\btheta)$ using without-replacement mini-batch SGD. 
We use $B$ as mini-batch size, so it takes $\epoch \defeq N/B$ steps to go over $N$ data points in the dataset. For simplicity we assume that $N$ is a multiple of $B$.
At iteration $t = k\epoch$, it partitions the dataset at random, into $\epoch$ sets of cardinality $B$: $B^{(k\epoch)}, B^{(k\epoch+1)}, \ldots, B^{(k\epoch+\epoch-1)}$, and uses these sets to estimate gradients. After each epoch (one pass through the dataset), the data is ``reshuffled'' and a new partition is used. Without-replacement SGD is known to be more difficult to analyze than with-replacement SGD (see \citep{shamir2016without, haochen2018random} and references therein), although more widely used in practice.

More concretely, our SGD algorithm uses the update rule
$\btheta^{(t+1)} \leftarrow \btheta^{(t)} - \eta g^{(t)}$,
where we fix the step size $\eta$ to be a constant throughout the entire run and $g^{(t)}$ is the gradient estimate 
\begin{equation*}
g^{(t)} = \tfrac{1}{B} \sum\nolimits_{i \in B^{(t)}} \ell'(f_{\btheta^{(t)}}(x_i);y_i) \nabla_\btheta f_{\btheta^{(t)}}(x_i).
\end{equation*}
For each $k$, $\bigcup_{t = k\epoch}^{k\epoch+\epoch-1} B^{(t)}= [N]$.
Note also that if $B = N$, we recover vanilla gradient descent.

Now consider a memorizing global minimum $\btheta^*$.
We define vectors $\nu_i \defeq \nabla_\btheta f_{\btheta^*}(x_i)$ for all $i \in [N]$.
We can then express any iterate $\btheta^{(t)}$ of SGD as
$\btheta^{(t)} = \btheta^* + \bxi^{(t)}$,
and then further decompose the ``perturbation'' $\bxi^{(t)}$ as the sum of two orthogonal components $\bxi_{\parallel}^{(t)}$ and $\bxi_{\perp}^{(t)}$,
where $\bxi_{\parallel}^{(t)} \in \spann(\{\nu_i\}_{i=1}^N)$ and $\bxi_{\perp}^{(t)} \in \spann(\{\nu_i\}_{i=1}^N)^{\perp}$.
Also, for a vector $v$, let $\norms{v}$ denote its $\ell_2$ norm.

\vspace*{-5pt}
\subsection{Main results and discussion}
\vspace*{-5pt}
We now state the main theorem of the section. For the proof, please refer to Appendix~\ref{sec:thmpf-memorizer}.
\begin{theorem}
\label{thm:memorizer}
Suppose a memorizing global minimum $\btheta^*$ of $\erisk(\btheta)$ is given,
and that $\erisk(\cdot)$ is differentiable at $\btheta^*$.
Then, there exist positive constants $\rho$, $\gamma$, $\lambda$, and $\tau$ satisfying the following:
if initialization $\btheta^{(0)}$ satisfies $\norms{\bxi^{(0)}} \leq \rho$, then 
\begin{align*}
\erisk(\btheta^{(0)})-\erisk(\btheta^*) = O(\norms{\bxi^{(0)}}^2),
\end{align*}
and SGD with step size $\eta < \gamma$ satisfies
\begin{align*}
\norms {\bxi_{\parallel}^{(kE+E)}} \leq (1-\eta \lambda) \norms {\bxi_{\parallel}^{(kE)}},~~\text{and}~~
\norms {\bxi^{(kE+E)}} \leq \norms {\bxi^{(kE)}} + \eta \lambda \norms {\bxi_{\parallel}^{(kE)}},
\end{align*}
as long as $\norms{\bxi_{\parallel}^{(t)}} \geq \tau \norms{\bxi^{(t)}}^2$ holds for all $t \in [kE, kE+E-1]$.
As a consequence, at the \textbf{first} iterate $t^* \geq 0$ where the condition $\norms{\bxi_{\parallel}^{(t)}} \geq \tau \norms{\bxi^{(t)}}^2$ is \textbf{violated}, we have
\begin{align*}
\norms {\bxi^{(t^*)}} \leq 2\norms {\bxi^{(0)}},~~\text{and}~~
\erisk(\btheta^{(t^*)})-\erisk(\btheta^*) \leq C \norms {\bxi^{(0)}}^4,
\end{align*}
for some positive constant $C$.
\end{theorem}

The full description of constants $\rho$, $\gamma$, $\lambda$, $\tau$, and $C$ can be found in Appendix~\ref{sec:thmpf-memorizer}. 
They are dependent on a number of terms, such as $N$, $B$, the Taylor expansions of loss $\ell(f_{\btheta^*}(x_i);y_i)$ and network output $f_{\btheta^*}(x_i)$ around the memorizing global minimum $\btheta^*$, maximum and minimum strictly positive eigenvalues of $H = \sum_{i=1}^N \ell''(f_{\btheta^*}(x_i);y_i) \nu_i \nu_i^T$.
The constant $\rho$ must be small enough so that as long as $\| \bxi \| \leq \rho$, the slopes of piecewise linear activation functions evaluated for data points $x_i$ do not change from $\btheta^*$ to $\btheta^* + \bxi$.

Notice that for small perturbation $\bxi$, the Taylor expansion of network output $f_{\btheta^*}(x_i)$ is written as
$f_{\btheta^*+\bxi}(x_i) = f_{\btheta^*}(x_i) + \nu_i ^T \bxi_{\parallel} + O(\norms{\bxi}^2)$,
because $\nu_i \perp \bxi_{\perp}$ by definition.
From this perspective, Theorem~\ref{thm:memorizer} shows that if initialized near global minima, the component in the perturbation $\bxi$ that induces first-order perturbation of $f_{\btheta^*}(x_i)$, namely $\bxi_{\parallel}$, decays exponentially fast until SGD finds a nearby point that has much smaller risk ($O(\norms {\bxi^{(0)}}^4)$) than the initialization ($O(\norms {\bxi^{(0)}}^2)$).
Note also that our result is completely deterministic, and independent of the partitions of the dataset taken by the algorithm; the theorem holds true even if the algorithm is not ``stochastic'' and just cycles through the dataset in a fixed order without reshuffling.

We would like to emphasize that Theorem~\ref{thm:memorizer} holds for \emph{any} memorizing global minima of FNNs, not only for the ones explicitly constructed in Sections~\ref{sec:fnns} and \ref{sec:genpos}. Moreover, the result is not dependent on the network size or data distribution. As long as the global minimum memorizes the data, our theorem holds \textit{without} any depth/width requirements or distributional assumptions, which is a noteworthy difference that makes our result hold in more realistic settings than existing ones.

The remaining question is: what happens after $t^*$? Unfortunately, if $\norms{\bxi_{\parallel}^{(t)}} \leq \tau \norms{\bxi^{(t)}}^2$, we cannot ensure exponential decay of $\norms{\bxi_{\parallel}^{(t)}}$, especially if it is small. Without exponential decay, one cannot show an upper bound on $\norms{\bxi^{(t)}}$ either. This means that after $t^*$, SGD may even diverge or oscillate near global minimum. Fully understanding the behavior of SGD after $t^*$ seems to be a more difficult problem, which we leave for future work.

\vspace*{-5pt}
\section{Conclusion and future work}
\vspace*{-5pt}
In this paper, we show that fully-connected neural networks (FNNs) with $\Omega(\sqrt N)$ nodes are expressive enough to perfectly memorize $N$ arbitrary data points, which is a significant improvement over the recent results in the literature.
We also prove the converse stating that at least $\Theta(\sqrt N)$ nodes are necessary; these two results together provide tight bounds on memorization capacity of neural networks.
We further extend our expressivity results to deeper and/or narrower networks, providing a nearly tight bound on memorization capacity for these networks as well. Under an assumption that data points are in general position, we prove that classification datasets can be memorized with $\Omega(N/d_x + d_y)$ hidden nodes in deep residual networks and one-hidden-layer FNNs, reducing the existing requirement of $\Omega(N)$. Finally, we study the dynamics of stochastic gradient descent (SGD) on empirical risk, and showed that if SGD is initialized near a global minimum that perfectly memorizes the data, it quickly finds a nearby point with small empirical risk. Several future topics are open; e.g., 1) tight bounds on memorization capacity for deep FNNs and other architectures,
2) deeper understanding of SGD dynamics in the presence of memorizing global minima.


\subsubsection*{Acknowledgments}
We thank Alexander Rakhlin for helpful discussion.
All the authors acknowledge support from DARPA Lagrange. Chulhee Yun also thanks Korea Foundation for Advanced Studies for their support. Suvrit Sra also acknowledges support from an NSF-CAREER grant and an Amazon Research Award.

\bibliographystyle{abbrvnat}
\bibliography{cite.bib}

\begin{thebibliography}{58}
\providecommand{\natexlab}[1]{#1}
\providecommand{\url}[1]{\texttt{#1}}
\expandafter\ifx\csname urlstyle\endcsname\relax
  \providecommand{\doi}[1]{doi: #1}\else
  \providecommand{\doi}{doi: \begingroup \urlstyle{rm}\Url}\fi

\bibitem[Allen-Zhu et~al.(2018)Allen-Zhu, Li, and Song]{allen2018convergence}
Z.~Allen-Zhu, Y.~Li, and Z.~Song.
\newblock A convergence theory for deep learning via over-parameterization.
\newblock \emph{arXiv preprint arXiv:1811.03962}, 2018.

\bibitem[Arpit et~al.(2017)Arpit, Jastrz{\k{e}}bski, Ballas, Krueger, Bengio,
  Kanwal, Maharaj, Fischer, Courville, Bengio, et~al.]{arpit2017closer}
D.~Arpit, S.~Jastrz{\k{e}}bski, N.~Ballas, D.~Krueger, E.~Bengio, M.~S. Kanwal,
  T.~Maharaj, A.~Fischer, A.~Courville, Y.~Bengio, et~al.
\newblock A closer look at memorization in deep networks.
\newblock In \emph{International Conference on Machine Learning}, pages
  233--242, 2017.

\bibitem[Bartlett et~al.(1999)Bartlett, Maiorov, and Meir]{bartlett1999almost}
P.~L. Bartlett, V.~Maiorov, and R.~Meir.
\newblock Almost linear {VC} dimension bounds for piecewise polynomial
  networks.
\newblock In \emph{Advances in Neural Information Processing Systems}, pages
  190--196, 1999.

\bibitem[Bartlett et~al.(2019{\natexlab{a}})Bartlett, Harvey, Liaw, and
  Mehrabian]{bartlett2019nearly}
P.~L. Bartlett, N.~Harvey, C.~Liaw, and A.~Mehrabian.
\newblock Nearly-tight {VC}-dimension and pseudodimension bounds for piecewise
  linear neural networks.
\newblock \emph{Journal of Machine Learning Research}, 20\penalty0
  (63):\penalty0 1--17, 2019{\natexlab{a}}.
\newblock URL \url{http://jmlr.org/papers/v20/17-612.html}.

\bibitem[Bartlett et~al.(2019{\natexlab{b}})Bartlett, Long, Lugosi, and
  Tsigler]{bartlett2019benign}
P.~L. Bartlett, P.~M. Long, G.~Lugosi, and A.~Tsigler.
\newblock Benign overfitting in linear regression.
\newblock \emph{arXiv preprint arXiv:1906.11300}, 2019{\natexlab{b}}.

\bibitem[Baum(1988)]{baum1988capabilities}
E.~B. Baum.
\newblock On the capabilities of multilayer perceptrons.
\newblock \emph{Journal of complexity}, 4\penalty0 (3):\penalty0 193--215,
  1988.

\bibitem[Belkin et~al.(2018{\natexlab{a}})Belkin, Hsu, Ma, and
  Mandal]{belkin2018reconciling}
M.~Belkin, D.~Hsu, S.~Ma, and S.~Mandal.
\newblock Reconciling modern machine learning and the bias-variance trade-off.
\newblock \emph{arXiv preprint arXiv:1812.11118}, 2018{\natexlab{a}}.

\bibitem[Belkin et~al.(2018{\natexlab{b}})Belkin, Rakhlin, and
  Tsybakov]{belkin2018does}
M.~Belkin, A.~Rakhlin, and A.~B. Tsybakov.
\newblock Does data interpolation contradict statistical optimality?
\newblock \emph{arXiv preprint arXiv:1806.09471}, 2018{\natexlab{b}}.

\bibitem[Brutzkus and Globerson(2017)]{brutzkus2017globally}
A.~Brutzkus and A.~Globerson.
\newblock Globally optimal gradient descent for a {ConvNet} with {G}aussian
  inputs.
\newblock In \emph{International Conference on Machine Learning}, pages
  605--614, 2017.

\bibitem[Brutzkus et~al.(2018)Brutzkus, Globerson, Malach, and
  Shalev-Shwartz]{brutzkus2018sgd}
A.~Brutzkus, A.~Globerson, E.~Malach, and S.~Shalev-Shwartz.
\newblock {SGD} learns over-parameterized networks that provably generalize on
  linearly separable data.
\newblock In \emph{International Conference on Learning Representations}, 2018.

\bibitem[Cover(1965)]{cover1965geometrical}
T.~M. Cover.
\newblock Geometrical and statistical properties of systems of linear
  inequalities with applications in pattern recognition.
\newblock \emph{IEEE transactions on electronic computers}, \penalty0
  (3):\penalty0 326--334, 1965.

\bibitem[Cybenko(1989)]{cybenko1989approximation}
G.~Cybenko.
\newblock Approximation by superpositions of a sigmoidal function.
\newblock \emph{Mathematics of control, signals and systems}, 2\penalty0
  (4):\penalty0 303--314, 1989.

\bibitem[Delalleau and Bengio(2011)]{delalleau2011shallow}
O.~Delalleau and Y.~Bengio.
\newblock Shallow vs. deep sum-product networks.
\newblock In \emph{Advances in Neural Information Processing Systems}, pages
  666--674, 2011.

\bibitem[Du et~al.(2017)Du, Lee, Tian, Poczos, and Singh]{du2017gradient}
S.~S. Du, J.~D. Lee, Y.~Tian, B.~Poczos, and A.~Singh.
\newblock Gradient descent learns one-hidden-layer {CNN}: Don't be afraid of
  spurious local minima.
\newblock \emph{arXiv preprint arXiv:1712.00779}, 2017.

\bibitem[Du et~al.(2018{\natexlab{a}})Du, Lee, Li, Wang, and
  Zhai]{du2018gradientB}
S.~S. Du, J.~D. Lee, H.~Li, L.~Wang, and X.~Zhai.
\newblock Gradient descent finds global minima of deep neural networks.
\newblock \emph{arXiv preprint arXiv:1811.03804}, 2018{\natexlab{a}}.

\bibitem[Du et~al.(2018{\natexlab{b}})Du, Zhai, Poczos, and
  Singh]{du2018gradientA}
S.~S. Du, X.~Zhai, B.~Poczos, and A.~Singh.
\newblock Gradient descent provably optimizes over-parameterized neural
  networks.
\newblock \emph{arXiv preprint arXiv:1810.02054}, 2018{\natexlab{b}}.

\bibitem[Eldan and Shamir(2016)]{eldan2016power}
R.~Eldan and O.~Shamir.
\newblock The power of depth for feedforward neural networks.
\newblock In \emph{Conference on Learning Theory}, pages 907--940, 2016.

\bibitem[Funahashi(1989)]{funahashi1989approximate}
K.-I. Funahashi.
\newblock On the approximate realization of continuous mappings by neural
  networks.
\newblock \emph{Neural networks}, 2\penalty0 (3):\penalty0 183--192, 1989.

\bibitem[HaoChen and Sra(2018)]{haochen2018random}
J.~Z. HaoChen and S.~Sra.
\newblock Random shuffling beats {SGD} after finite epochs.
\newblock \emph{arXiv preprint arXiv:1806.10077}, 2018.

\bibitem[Hardt and Ma(2017)]{hardt2017identity}
M.~Hardt and T.~Ma.
\newblock Identity matters in deep learning.
\newblock In \emph{International Conference on Learning Representations}, 2017.

\bibitem[Hornik et~al.(1989)Hornik, Stinchcombe, and
  White]{hornik1989multilayer}
K.~Hornik, M.~Stinchcombe, and H.~White.
\newblock Multilayer feedforward networks are universal approximators.
\newblock \emph{Neural networks}, 2\penalty0 (5):\penalty0 359--366, 1989.

\bibitem[Huang(2003)]{huang2003learning}
G.-B. Huang.
\newblock Learning capability and storage capacity of two-hidden-layer
  feedforward networks.
\newblock \emph{IEEE Transactions on Neural Networks}, 14\penalty0
  (2):\penalty0 274--281, 2003.

\bibitem[Huang and Babri(1998)]{huang1998upper}
G.-B. Huang and H.~A. Babri.
\newblock Upper bounds on the number of hidden neurons in feedforward networks
  with arbitrary bounded nonlinear activation functions.
\newblock \emph{IEEE Transactions on Neural Networks}, 9\penalty0 (1):\penalty0
  224--229, 1998.

\bibitem[Huang and Huang(1991)]{huang1991bounds}
S.-C. Huang and Y.-F. Huang.
\newblock Bounds on the number of hidden neurons in multilayer perceptrons.
\newblock \emph{IEEE transactions on neural networks}, 2\penalty0 (1):\penalty0
  47--55, 1991.

\bibitem[Kawaguchi(2016)]{kawaguchi2016deep}
K.~Kawaguchi.
\newblock Deep learning without poor local minima.
\newblock In \emph{Advances in Neural Information Processing Systems}, pages
  586--594, 2016.

\bibitem[Kowalczyk(1997)]{kowalczyk1997estimates}
A.~Kowalczyk.
\newblock Estimates of storage capacity of multilayer perceptron with threshold
  logic hidden units.
\newblock \emph{Neural networks}, 10\penalty0 (8):\penalty0 1417--1433, 1997.

\bibitem[Laurent and Brecht(2018)]{laurent2018deep}
T.~Laurent and J.~Brecht.
\newblock Deep linear networks with arbitrary loss: All local minima are
  global.
\newblock In \emph{International Conference on Machine Learning}, pages
  2908--2913, 2018.

\bibitem[Li and Liang(2018)]{li2018learning}
Y.~Li and Y.~Liang.
\newblock Learning overparameterized neural networks via stochastic gradient
  descent on structured data.
\newblock In \emph{Advances in Neural Information Processing Systems}, pages
  8168--8177, 2018.

\bibitem[Li and Yuan(2017)]{li2017convergence}
Y.~Li and Y.~Yuan.
\newblock Convergence analysis of two-layer neural networks with {ReLU}
  activation.
\newblock In \emph{Advances in Neural Information Processing Systems}, pages
  597--607, 2017.

\bibitem[Liang and Srikant(2017)]{liang2017deep}
S.~Liang and R.~Srikant.
\newblock Why deep neural networks for function approximation?
\newblock In \emph{International Conference on Learning Representations}, 2017.

\bibitem[Liang and Rakhlin(2018)]{liang2018just}
T.~Liang and A.~Rakhlin.
\newblock {Just Interpolate: Kernel ``Ridgeless'' Regression Can Generalize}.
\newblock \emph{arXiv preprint arXiv:1808.00387}, 2018.

\bibitem[Liang et~al.(2019)Liang, Rakhlin, and Zhai]{liang2019risk}
T.~Liang, A.~Rakhlin, and X.~Zhai.
\newblock On the risk of minimum-norm interpolants and restricted lower
  isometry of kernels.
\newblock \emph{arXiv preprint arXiv:1908.10292}, 2019.

\bibitem[Lu et~al.(2017)Lu, Pu, Wang, Hu, and Wang]{lu2017expressive}
Z.~Lu, H.~Pu, F.~Wang, Z.~Hu, and L.~Wang.
\newblock The expressive power of neural networks: A view from the width.
\newblock In \emph{Advances in Neural Information Processing Systems}, pages
  6231--6239, 2017.

\bibitem[Mei and Montanari(2019)]{mei2019generalization}
S.~Mei and A.~Montanari.
\newblock The generalization error of random features regression: Precise
  asymptotics and double descent curve.
\newblock \emph{arXiv preprint arXiv:1908.05355}, 2019.

\bibitem[Nguyen and Hein(2017)]{nguyen2017losscnn}
Q.~Nguyen and M.~Hein.
\newblock Optimization landscape and expressivity of deep {CNN}s.
\newblock \emph{arXiv preprint arXiv:1710.10928}, 2017.

\bibitem[Nilsson(1965)]{nilsson1965learning}
N.~J. Nilsson.
\newblock Learning machines.
\newblock 1965.

\bibitem[Rolnick and Tegmark(2018)]{rolnick2018power}
D.~Rolnick and M.~Tegmark.
\newblock The power of deeper networks for expressing natural functions.
\newblock In \emph{International Conference on Learning Representations}, 2018.

\bibitem[Safran and Shamir(2017{\natexlab{a}})]{safran2017depth}
I.~Safran and O.~Shamir.
\newblock Depth-width tradeoffs in approximating natural functions with neural
  networks.
\newblock In \emph{International Conference on Machine Learning}, pages
  2979--2987, 2017{\natexlab{a}}.

\bibitem[Safran and Shamir(2017{\natexlab{b}})]{safran2017spurious}
I.~Safran and O.~Shamir.
\newblock Spurious local minima are common in two-layer {ReLU} neural networks.
\newblock \emph{arXiv preprint arXiv:1712.08968}, 2017{\natexlab{b}}.

\bibitem[Shamir(2016)]{shamir2016without}
O.~Shamir.
\newblock Without-replacement sampling for stochastic gradient methods.
\newblock In \emph{Advances in neural information processing systems}, pages
  46--54, 2016.

\bibitem[Soltanolkotabi(2017)]{soltanolkotabi2017learning}
M.~Soltanolkotabi.
\newblock Learning {ReLUs} via gradient descent.
\newblock In \emph{Advances in Neural Information Processing Systems}, pages
  2007--2017, 2017.

\bibitem[Sontag(1997)]{sontag1997shattering}
E.~D. Sontag.
\newblock Shattering all sets of `k' points in ``general position'' requires
  (k—1)/2 parameters.
\newblock \emph{Neural Computation}, 9\penalty0 (2):\penalty0 337--348, 1997.

\bibitem[Soudry and Carmon(2016)]{soudry2016no}
D.~Soudry and Y.~Carmon.
\newblock No bad local minima: Data independent training error guarantees for
  multilayer neural networks.
\newblock \emph{arXiv preprint arXiv:1605.08361}, 2016.

\bibitem[Telgarsky(2015)]{telgarsky2015representation}
M.~Telgarsky.
\newblock Representation benefits of deep feedforward networks.
\newblock \emph{arXiv preprint arXiv:1509.08101}, 2015.

\bibitem[Telgarsky(2016)]{telgarsky2016benefits}
M.~Telgarsky.
\newblock Benefits of depth in neural networks.
\newblock In \emph{Conference on Learning Theory}, pages 1517--1539, 2016.

\bibitem[Tian(2017)]{tian2017analytical}
Y.~Tian.
\newblock An analytical formula of population gradient for two-layered {ReLU}
  network and its applications in convergence and critical point analysis.
\newblock In \emph{International Conference on Machine Learning}, pages
  3404--3413, 2017.

\bibitem[Wang et~al.(2018)Wang, Giannakis, and Chen]{wang2018learning}
G.~Wang, G.~B. Giannakis, and J.~Chen.
\newblock Learning {ReLU} networks on linearly separable data: Algorithm,
  optimality, and generalization.
\newblock \emph{arXiv preprint arXiv:1808.04685}, 2018.

\bibitem[Yamasaki(1993)]{yamasaki1993lower}
M.~Yamasaki.
\newblock The lower bound of the capacity for a neural network with multiple
  hidden layers.
\newblock In \emph{ICANN'93}, pages 546--549. Springer, 1993.

\bibitem[Yarotsky(2017)]{yarotsky2017error}
D.~Yarotsky.
\newblock Error bounds for approximations with deep {ReLU} networks.
\newblock \emph{Neural Networks}, 94:\penalty0 103--114, 2017.

\bibitem[Yarotsky(2018)]{yarotsky2018optimal}
D.~Yarotsky.
\newblock Optimal approximation of continuous functions by very deep {ReLU}
  networks.
\newblock \emph{arXiv preprint arXiv:1802.03620}, 2018.

\bibitem[Yun et~al.(2018)Yun, Sra, and Jadbabaie]{yun2018global}
C.~Yun, S.~Sra, and A.~Jadbabaie.
\newblock Global optimality conditions for deep neural networks.
\newblock In \emph{International Conference on Learning Representations}, 2018.

\bibitem[Yun et~al.(2019)Yun, Sra, and Jadbabaie]{yun2019small}
C.~Yun, S.~Sra, and A.~Jadbabaie.
\newblock Small nonlinearities in activation functions create bad local minima
  in neural networks.
\newblock In \emph{International Conference on Learning Representations}, 2019.

\bibitem[Zhang et~al.(2017)Zhang, Bengio, Hardt, Recht, and
  Vinyals]{zhang2017understanding}
C.~Zhang, S.~Bengio, M.~Hardt, B.~Recht, and O.~Vinyals.
\newblock Understanding deep learning requires rethinking generalization.
\newblock In \emph{International Conference on Learning Representations
  (ICLR)}, 2017.

\bibitem[Zhang et~al.(2018)Zhang, Yu, Wang, and Gu]{zhang2018learning}
X.~Zhang, Y.~Yu, L.~Wang, and Q.~Gu.
\newblock Learning one-hidden-layer {ReLU} networks via gradient descent.
\newblock \emph{arXiv preprint arXiv:1806.07808}, 2018.

\bibitem[Zhong et~al.(2017)Zhong, Song, Jain, Bartlett, and
  Dhillon]{zhong2017recovery}
K.~Zhong, Z.~Song, P.~Jain, P.~L. Bartlett, and I.~S. Dhillon.
\newblock Recovery guarantees for one-hidden-layer neural networks.
\newblock In \emph{International Conference on Machine Learning}, pages
  4140--4149, 2017.

\bibitem[Zhou and Liang(2018)]{zhou2018critical}
Y.~Zhou and Y.~Liang.
\newblock Critical points of neural networks: Analytical forms and landscape
  properties.
\newblock In \emph{International Conference on Learning Representations}, 2018.

\bibitem[Zhou et~al.(2019)Zhou, Yang, Zhang, Liang, and Tarokh]{zhou2019sgd}
Y.~Zhou, J.~Yang, H.~Zhang, Y.~Liang, and V.~Tarokh.
\newblock {SGD} converges to global minimum in deep learning via star-convex
  path.
\newblock In \emph{International Conference on Learning Representations}, 2019.

\bibitem[Zou et~al.(2018)Zou, Cao, Zhou, and Gu]{zou2018stochastic}
D.~Zou, Y.~Cao, D.~Zhou, and Q.~Gu.
\newblock Stochastic gradient descent optimizes over-parameterized deep {ReLU}
  networks.
\newblock \emph{arXiv preprint arXiv:1811.08888}, 2018.

\end{thebibliography}

\newpage
\clearpage
\appendix

\vspace*{-5pt}
\section{Deferred theorem statements}
\vspace*{-5pt}
\label{sec:thm-deferred}
In this section, we state the theorems that were omitted in Section~\ref{sec:deeper} due to lack of space. First, we start by stating the ReLU-like version of Theorem~\ref{thm:htanml}: 
\begin{corollary}
	\label{cor:reluml}
	Consider any dataset $\{(x_i, y_i)\}_{i=1}^N$ that satisfies Assumption~\ref{asm:data}. 
	For an $L$-layer FNN with ReLU(-like) activation \textup{($\relulikeact$)}, assume that there exist indices $l_1, \dots, l_m \in [L-2]$ that satisfies
	\begin{itemize}
		\vspace*{-5pt}
		\setlength{\itemsep}{0pt}
		\item $l_j + 1 < l_{j+1}$ for $j \in [m-1]$,
		\item $4 \sum_{j=1}^{m} \left \lfloor \frac{d_{l_j}-r_j}{4} \right \rfloor \left \lfloor \frac{d_{l_j+1}-r_j}{4d_y} \right \rfloor \geq N$, where $r_j = d_y \indic{j>1}+\indic{j<m}$, for $j \in [m]$,
		\item $d_k \geq d_y + 1$ for all $k \in \bigcup_{j \in [m-1]}[l_j + 2:l_{j+1}-1]$.
		\item $d_k \geq d_y$ for all $k \in [l_m + 2:L-1]$,
		\vspace*{-5pt}
	\end{itemize}
	where $\indic{\cdot}$ is 0-1 indicator function. Then, there exists $\btheta$ such that $y_i = f_{\btheta} (x_i)$ for all $i \in [N]$.
\end{corollary}
The idea is that anything that holds for hard-tanh activation holds for ReLU networks that has double the width. One difference to note is that the number of nodes needed for ``propagating'' input and output information (the circle and diamond nodes in Figure~\ref{fig:fig2}) has not doubled. This is because merely propagating the information without nonlinear distortion can be done with a single ReLU-like activation.

The next corollaries are special cases for classification. One can check that with $L = 4$ and $m = 2$ (hence $l_1 = 1$ and $l_2 = 3$), these boil down to Proposition~\ref{prop:htanclsf4l}.
\begin{corollary}
	\label{cor:htanclsfml}
	Consider any dataset $\{(x_i, y_i)\}_{i=1}^N$ that satisfies Assumption~\ref{asm:data}. 
	Assume that $y_i \in \{0,1\}^{d_y}$ is the one-hot encoding of $d_y$ classes.
	For an $L$-layer FNN with hard-tanh activation \textup{($\htanhact$)}, assume that there exist indices $l_1, \dots, l_m \in [L-1]$ \textup{($m\geq2$)} that satisfies
	\begin{itemize}
		\vspace*{-5pt}
		\setlength{\itemsep}{0pt}
		\item $l_j + 1 < l_{j+1}$ for $j \in [m-1]$,
		\item $4 \sum\limits_{j=1}^{m-1} \left \lfloor \frac{d_{l_j}-r_j}{2} \right \rfloor \left  \lfloor \frac{d_{l_j+1}-r_j}{2} \right \rfloor \geq N$, where $r_j = \indic{j>1}+\indic{j<m-1}$, for $j \in [m-1]$,
		\item $d_{l_m} \geq 2 d_y$,
		\item $d_k \geq 2$ for all $k \in \bigcup_{j \in [m-2]}[l_j + 2:l_{j+1}-1]$.
		\item $d_k \geq d_y$ for all $k \in [l_m + 1: L-1]$.
		\vspace*{-5pt}
	\end{itemize}
	Then, there exists $\btheta$ such that $y_i = f_{\btheta} (x_i)$ for all $i \in [N]$.
\end{corollary}

\begin{corollary}
	\label{cor:reluclsfml}
	Consider any dataset $\{(x_i, y_i)\}_{i=1}^N$ that satisfies Assumption~\ref{asm:data}. 
	Assume that $y_i \in \{0,1\}^{d_y}$ is the one-hot encoding of $d_y$ classes.
	For an $L$-layer FNN with ReLU(-like) activation \textup{($\relulikeact$)}, assume that there exist indices $l_1, \dots, l_m \in [L-1]$ \textup{($m\geq2$)} that satisfies
	\begin{itemize}
		\vspace*{-5pt}
		\setlength{\itemsep}{0pt}
		\item $l_j + 1 < l_{j+1}$ for $j \in [m-1]$,
		\item $4 \sum\limits_{j=1}^{m-1} \left \lfloor \frac{d_{l_j}-r_j}{4} \right \rfloor \left \lfloor \frac{d_{l_j+1}-r_j}{4} \right \rfloor \geq N$, where $r_j = \indic{j>1}+\indic{j<m-1}$, for $j \in [m-1]$,
		\item $d_{l_m} \geq 4 d_y$,
		\item $d_k \geq 2$ for all $k \in \bigcup_{j \in [m-2]}[l_j + 2:l_{j+1}-1]$.
		\item $d_k \geq d_y$ for all $k \in [l_m + 1: L-1]$.
		\vspace*{-5pt}
	\end{itemize}
	Then, there exists $\btheta$ such that $y_i = f_{\btheta} (x_i)$ for all $i \in [N]$.
\end{corollary}
The proof of Corollaries~\ref{cor:htanclsfml} and \ref{cor:reluclsfml} can be done by easily combining the ideas in proofs of Proposition~\ref{prop:htanclsf4l} and Proposition~\ref{thm:htanml}, hence omitted.

\vspace*{-5pt}
\section{Proof of Theorem~\ref{thm:htan3l}}
\vspace*{-5pt}
\label{sec:thmpf-htan3l}
We prove the theorem by constructing a parameter $\btheta$ that perfectly fits the dataset.
We will prove the theorem for hard-tanh ($\htanhact$) only, because extension to ReLU-like ($\relulikeact$) is straightforward from its definition.
To convey the main idea more clearly, we first prove the theorem for $d_y = 1$, and later discuss how to extend to $d_y > 1$.

For a data point $x_i$, the corresponding input and output of the $l$-th hidden layer is written as $z^{l}(x_i)$ and $a^{l}(x_i)$, respectively. Moreover, $z_j^{l}(x_i)$ and $a_j^{l}(x_i)$ denote the input and output of the $j$-th node of the $l$-th hidden layer.
For weight matrices $\bW^{l}$, we will denote its $(j,k)$-th entry as $\Wmatent{l}{j}{k}$, its $j$-th row as $\Wmatrow{l}{j}$, and its $j$-th column as $\bW^l_{:,j}$. Similarly, $\bveccomp{l}{j}$ denote the $j$-th component of the bias vector $\bb^{l}$.
To simplify notation, we will denote $p \defeq d_1$ and $q \defeq d_2$, for the rest of the proof. Assume for simplicity that $p$ is a multiple of $2$, $q$ is a multiple of $2$, and $pq = N$.

\vspace*{-5pt}
\subsection{Proof sketch}
\vspace*{-5pt}
The proof consists of three steps, one for each layer. In this subsection, we will describe each step in the following three paragraphs. Then, the next three subsections will provide the full details of each step.

In the first step, we down-project all input data points to a line, using a random vector $u \in \reals^{d_x}$. Different $x_i$'s are mapped to different $u^T x_i$'s, so we have $N$ distinct $u^T x_i$'s on the line. Now re-index the data points in increasing order of $u^T x_i$, and divide total $N$ data points into $p$ groups with $q$ points each.
To do this, each row $\Wmatrow{1}{j}$ of $\bW^{1}$ is chosen as $u^T$ multiplied by a scalar.
We choose the appropriate scalar for $\Wmatrow{1}{j}$ and bias $\bveccomp{1}{j}$, so that the input to the $j$-th hidden node in layer~1, $z_j^{1}(\cdot)$, satisfies the following: (1)~$z_j^{1}(x_i) \in (-1,1)$ for indices $i \in [jq-q+1:jq]$, and (2) $z_j^{1}(x_i) \in (-1,1)^c$ for all other indices so that they are ``clipped'' by $\htanhact$.

\begin{figure}[t]
	\centering
	\includegraphics[width=0.96\linewidth]{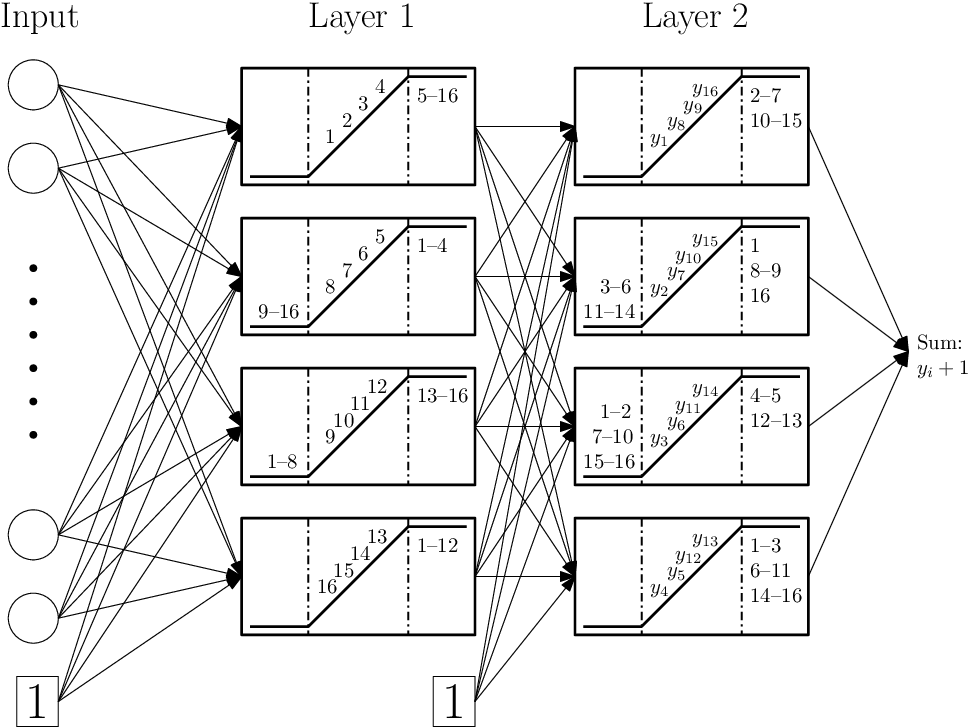}
	\vspace*{18pt}
	\caption{Illustration of the construction for $d_1 = d_2 =4$. Each box corresponds to a hidden node with hard-tanh activation. In each hidden node, the numbers written in the three parts are indices of data points that are clipped to $-1$ at output (left), those clipped to $+1$ (right), and those unchanged (center). One can check for all indices that outputs of layer~2 sum to $y_i + 1$.}
	\label{fig:fig1}
\end{figure}

In the second step, for each hidden node in layer~2, we pick one point each from these $p$ groups and map their values to desired $y_i$. More specifically, for $k$-th node in layer~2, we define an index set $\mc I_k$ (with cardinality $p$) that contains exactly one element from each $[jq-q+1:jq]$, and choose $\Wmatrow{2}{k}$ and $\bveccomp{2}{k}$ such that $z_k^{2}(x_i) = y_i$ for $i \in \mc I_k$ and $z_k^{2}(x_i) \in [-1,1]^c$ for $i \notin \mc I_k$. This is possible because for each $k$, we are solving $p$ linear equations with $p+1$ variables.

As we will see in the details, the first and second steps involve alternating signs and a carefully designed choice of index sets $\mc I_k$ so that sum of output $a_k^{2}(\cdot)$ of each node in layer~2 becomes $y_i+1$. Figure~\ref{fig:fig1} shows a simple illustration for $p = q = 4$.
With this choice, we can make the output $f_\btheta(x_i)$ become simply $y_i$ for all $i \in [N]$, thereby perfectly memorizing the dataset. 

\vspace*{-5pt}
\subsection{Input to layer~1: down-project and divide}
\vspace*{-5pt}
\label{sec:pf-inp-l1}
First, recall from Assumption~\ref{asm:data} that all $x_i$'s are distinct. This means that for any pair of data points $x_i$ and $x_{i'}$, the set of vectors $u \in \reals^{d_x}$ satisfying $u^T x_i = u^T x_{i'}$ has measure zero. Thus, if we sample any $u$ from some distribution (e.g., Gaussian), $u$ satisfies $u^T x_i \neq u^T x_{i'}$ for all $i \neq i'$ with probability~1. This is a standard proof technique also used in other papers; please see e.g., \citet[Lemma~2.1]{huang2003learning}.

We choose any such $u$, and without loss of generality, re-index the data points in increasing order of $u^T x_i$: $u^T x_1 < u^T x_2 < \cdots < u^T x_N$. 
Now define $c_i \defeq u^T x_i$ for all $i \in [N]$, and additionally, $c_0 = c_1 - \delta$ and $c_{N+1} = c_N + \delta$, for any $\delta > 0$.

Now, we are going to define $\bW^{1}$ and $\bb^{1}$ such that the input to the $j$-th ($j \in [p]$) hidden node in layer~1 has $z_j^{1}(x_i) \in (-1,1)$ for indices $i \in [jq-q+1:jq]$, and $z_j^{1}(x_i) \in (-1,1)^c$ for any other points.
We also alternate the order of data points, which will prove useful in later steps.
More concretely, we define the $j$-th row of $\bW^{1}$ and $j$-th component of $\bb^{1}$ to be
\begin{align*}
\Wmatrow{1}{j} &= (-1)^{j-1} \frac{4}{c_{jq}+c_{jq+1}-c_{jq-q}-c_{jq-q+1}}u^T,\\
\bveccomp{1}{j} &= (-1)^{j} \frac{c_{jq}+c_{jq+1}+c_{jq-q}+c_{jq-q+1}}{c_{jq}+c_{jq+1}-c_{jq-q}-c_{jq-q+1}}.
\end{align*}
When $j$ is odd, it is easy to check that $z_j^{1}(\cdot)$ satisfies 
\begin{align*}
&-1 < z_j^{1}(x_{jq-q+1})  < \cdots < z_j^{1}(x_{jq}) < +1,\\
&z_j^{1} (x_{i}) < -1 \text{ for } i \leq jq-q,\\
&z_j^{1}(x_{i}) > +1 \text{ for } i > jq,
\end{align*}
so that the output $a_j^{1}(\cdot)$ satisfies
\begin{align}
&-1 < a_j^{1}(x_{jq-q+1})  < \cdots < a_j^{1}(x_{jq}) < +1, \label{eq:st2midodd}\\
&a_j^{1} (x_{i}) = -1 \text{ for } i \leq jq-q,\label{eq:st2clipoddless}\\
&a_j^{1} (x_{i}) = +1 \text{ for } i > jq.\label{eq:st2clipoddmore}
\end{align}
When $j$ is even, by a similar argument:
\begin{align}
&+1 > a_j^{1}(x_{jq-q+1}) > \cdots > a_j^{1}(x_{jq}) > -1, \label{eq:st2mideven} \\
&a_j^{1} (x_{i}) = +1 \text{ for } i \leq jq-q,\label{eq:st2clipevenless}\\
&a_j^{1}(x_{i}) = -1 \text{ for } i > jq. \label{eq:st2clipevenmore}
\end{align}

\vspace*{-5pt}
\subsection{Layer~1 to 2: place at desired positions}
\vspace*{-5pt}
At each node of layer~2, we will show how to place $p$ points at the right position, and the rest of points in the clipping region.
After that, we will see that adding up all node outputs of layer~2 gives $y_i+1$ for all $i$.

For $k$-th hidden node in layer~2 ($k \in [q]$), define a set 
\begin{equation*}
\mc I_k \defeq 
\{k, 2q+1-k, 2q+k, 4q+1-k, \dots, pq+1-k\}.
\end{equation*}
Note that $|\mc I_k| = p$. Also, let us denote the elements of $\mc I_k$ as $i_{k,1}, \dots, i_{k,p}$ in increasing order. For example, $i_{k,1} = k$, $i_{k,2} = 2q+1-k$, and so on. We can see that $i_{k,j} \in [jq-q+1:jq]$.

For each $k$, our goal is to construct $\Wmatrow{2}{k}$ and $\bveccomp{2}{k}$ so that the input to the $k$-th node of layer~2 places data points indexed with $i \in \mc I_k$ to the desired position $y_i \in [-1,1]$, and the rest of data points $i \notin \mc I_k$ outside $[-1,1]$.

\vspace*{-5pt}
\paragraph{Case 1: odd $k$.}
We first describe how to construct $\Wmatrow{2}{k}$ and $\bveccomp{2}{k}$ for \textbf{odd} $k$'s.
First of all, consider data points $x_{i_{k,j}}$'s in $\mc I_k$. We want to choose parameters so that the input to the $k$-th node is equal to $y_{i_{k,j}}$'s:
\begin{align*}
&z_k^{2}(x_{i_{k,j}}) =
\sum\nolimits_{l=1}^p \Wmatent{2}{k}{l} a_l^{1}(x_{i_{k,j}})
+ \bveccomp{2}{k}
= y_{i_{k,j}},
\end{align*}
for all $j \in [p]$.
This is a system of $p$ linear equations with $p+1$ variables, which can be represented in a matrix-vector product form:
\begin{equation}
\label{eq:syslineq}
M_k
\begin{bmatrix}
(\Wmatrow{2}{k})^T \\
\bveccomp{2}{k}
\end{bmatrix}
=
\begin{bmatrix}
y_{i_{k,1}}\\
\vdots \\
y_{i_{k,p}}
\end{bmatrix},
\end{equation}
where the $(j,l)$-th entry of matrix $M_k \in \reals^{p \times (p+1)}$ is defined by $a_l^{1}(x_{i_{k,j}})$ for $j \in [p]$ and $l \in [p]$, and $(j,p+1)$-th entries are all equal to 1.

With the matrix $M_k$ defined from the above equation, we state the lemma whose simple proof is deferred to Appendix~\ref{sec:lemst3matrix} for better readability:
\begin{lemma}
	\label{lem:st3matrix}
	For any $k \in [q]$, the matrix $M_k \in \reals^{p \times (p+1)}$ satisfies the following properties:
	\begin{enumerate}
		\vspace*{-8pt}
		\setlength{\itemsep}{0pt}
		\item $M_k$ has full column rank.
		\item There exists a vector $\nu \in \nulsp(M_k)$ such that the first $p$ components of $\nu$ are all strictly positive.
		\vspace*{-8pt}
	\end{enumerate}
\end{lemma}
Lemma~\ref{lem:st3matrix} implies that for any $y_{i_{k,1}}, \dots, y_{i_{k,p}}$, there exist infinitely many solutions $(\Wmatrow{2}{k},\bveccomp{2}{k})$ for \eqref{eq:syslineq} of the form $\mu + \alpha \nu$, where $\mu$ is any particular solution satisfying the linear system and $\alpha$ is any scalar. This means that by scaling $\alpha$, and we can make $\Wmatrow{2}{k}$ as large as we want, without hurting $z_k^{2}(x_i) = y_i$ for $i \in \mc I_k$.

It is now left to make sure that any other data points $i \notin \mc I_k$ have $z_k^{2}(x_i) \in [-1,1]^c$. As we will show, this can be done by making $\alpha > 0$ sufficiently large.

Now fix any odd $j \in [p]$, and consider $i_{k,j} \in \mc I_k$, and recall $i_{k,j} \in [jq-q+1:jq]$.
Fix any other $i \in [jq-q+1:i_{k,j}-1]$. By Eqs \eqref{eq:st2clipoddless}, \eqref{eq:st2clipoddmore}, \eqref{eq:st2clipevenless} and \eqref{eq:st2clipevenmore}, 
the output of $l$-th node in layer~1 ($l \neq j$) is the same for $i$ and $i_{k,j}$: 
$a_{l}^{1}(x_i) = a_{l}^{1}(x_{i_{k,j}})$.

In contrast, for $a_{j}^{1}(\cdot)$, we have $a_j^{1}(x_{i}) < a_j^{1}(x_{i_{k,j}})$ \eqref{eq:st2midodd}.
Since $z_k^{2}(x_{i_{k,j}}) = \sum_{l} \Wmatent{2}{k}{l} a_l^{1}(x_{i_{k,j}})
+ \bveccomp{2}{k} = y_{i_{k,j}}$, large enough $\Wmatent{2}{k}{j}>0$ will make $z_k^{2}(x_i) < -1$, resulting in $a_k^{2}(x_i) = -1$; the output for $x_i$ is clipped.
A similar argument can be repeated for $i \in [i_{k,j}+1:jq]$, so that for large enough $\Wmatent{2}{k}{j}>0$,
\begin{align*}
a_k^{2}(x_{i}) &= -1,~\forall i \in [jq-q+1:i_{k,j}-1]\\
a_k^{2}(x_{i}) &= +1,~\forall i \in [i_{k,j}+1:jq].
\end{align*}
Similarly, for even $j \in [p]$, large $\Wmatent{2}{k}{j}>0$ will make 
\begin{align*}
a_k^{2}(x_{i}) &= +1,~\forall i \in [jq-q+1:i_{k,j}-1]\\
a_k^{2}(x_{i}) &= -1,~\forall i \in [i_{k,j}+1:jq].
\end{align*}

Summarizing, for large enough $\Wmatrow{2}{k}>0$ (achieved by making $\alpha > 0$ large), 
the output of the $k$-th node of layer~2 satisfies $a_k^{2}(x_i) = y_i,~\forall i \in \mc I_k$, and
\begin{align*}
&a_k^{2}(x_i) = -1,~\forall i \in \bigcup\nolimits_{\substack{j \in [0:p] \\ j \text{ even}}} [i_{k,j}+1:i_{k,j+1}-1],
\numberthis \label{eq:st3clipoddm}\\
&a_k^{2}(x_i) = +1,~\forall i \in \bigcup\nolimits_{\substack{j \in [p] \\ j \text{ odd}}} [i_{k,j}+1:i_{k,j+1}-1], \numberthis \label{eq:st3clipoddp}
\end{align*}
where $i_{k,0} \defeq 0$ and $i_{k,p+1} \defeq N+1$ for all $k \in [q]$.

\vspace*{-5pt}
\paragraph{Case 2: even $k$.}
For even $k$'s, we can repeat the same process, except that we push $\alpha<0$ to large negative number, so that $\Wmatrow{2}{k}<0$ is sufficiently large negative. By following a very similar argument, we can make the output of the $k$-th node of layer~2 satisfy $a_k^{2}(x_i) = y_i,~\forall i \in \mc I_k$, and 
\begin{align*}
&a_k^{2}(x_i) = +1,~\forall i \in \bigcup\nolimits_{\substack{j \in [0:p] \\ j \text{ even}}} [i_{k,j}+1\!:\!i_{k,j+1}-1], \numberthis \label{eq:st3clipevenp} \\
&a_k^{2}(x_i) = -1,~\forall i \in \bigcup\nolimits_{\substack{j \in [p] \\ j \text{ odd}}} [i_{k,j}+1:i_{k,j+1}-1]. \numberthis \label{eq:st3clipevenm}
\end{align*}

\vspace*{-5pt}
\subsection{Layer~2 to output: add them all}
\vspace*{-5pt}
Quite surprisingly, adding up $a_k^{2}(x_i)$ for all $k \in [q]$ gives $y_i+1$ for all $i \in [N]$. 
To prove this, first observe that the index sets $\mc I_1, \mc I_2, \dots, \mc I_q$ form a partition of $[N]$. So, proving $\sum_{l=1}^q a_l^{2}(x_{i_{k,j}}) = y_{i_{k,j}} + 1$ for all $j \in [p]$ and $k \in [q]$ suffices.

By the definition of $i_{k,1} = k, i_{k,2} = 2q+1-k, i_{k,3} = 2q+k, \dots, i_{k,p-1} = (p-2)q+k, i_{k,p} = pq + 1 - k$, we can see the following chains of inequalities:
\begin{align*}
&jq - q + 1= i_{1,j} < i_{2,j} < \cdots < i_{q,j} = jq ~\text { for $j$ odd,}\\
&jq - q + 1= i_{q,j} < \cdots < i_{2,j} < i_{1,j} = jq ~\text { for $j$ even.}
\end{align*}

Fix any $k \in [q]$, and any odd $j \in [p]$. From the above chains of inequalities, we can observe that
\begin{align*}
i_{k,j} &\in [i_{l,j}+1:i_{l,j+1}-1] ~\text{ if } l < k,\\
i_{k,j} &\in [i_{l,j-1}+1:i_{l,j}-1] ~\text{ if } l > k.
\end{align*}
Now, for $x_{i_{k,j}}$, we will sum up $a_l^{2}(x_{i_{k,j}})$ for $l \in [q]$.
First, for $1 \leq l < k$, we have $i_{k,j} \in [i_{l,j}+1:i_{l,j+1}-1]$. Since $j$ is odd, from Eqs \eqref{eq:st3clipoddp} and \eqref{eq:st3clipevenm},
\begin{equation*}
a_l^{2}(x_{i_{k,j}}) = \begin{cases}
+1 & \text{ for odd } l < k,\\
-1 & \text{ for even } l < k.\\
\end{cases}
\end{equation*} 
Similarly, for $k < l \leq w$, we have $i_{k,j} \in [i_{l,j-1}+1:i_{l,j}-1]$. Since $j$ is odd, from 
Eqs~\eqref{eq:st3clipoddm} and \eqref{eq:st3clipevenp},
\begin{equation*}
a_l^{2}(x_{i_{k,j}}) = \begin{cases}
-1 & \text{ for odd } l > k,\\
+1 & \text{ for even } l > k.\\
\end{cases}
\end{equation*}
Then, the sum over $l \neq k$ always results in $+1$, so
\begin{equation*}
\sum\nolimits_{l=1}^q a_l^{2}(x_{i_{k,j}}) 
= y_{i_{k,j}} + \sum\nolimits_{l \neq k} a_l^{2}(x_{i_{k,j}})
= y_{i_{k,j}} + 1.
\end{equation*}

For any fixed even $j \in [p]$, we can similarly prove the same thing. 
We have
\begin{align*}
i_{k,j} &\in [i_{l,j-1}+1: i_{l,j}-1] ~\text{ if } l < k,\\
i_{k,j} &\in [i_{l,j}+1: i_{l,j+1}-1] ~\text{ if } l > k,
\end{align*}
for even $j$. From this point, the remaining steps are exactly identical to the odd case.

Now that we know $\sum_{l=1}^q a_l^{2}(x_i) = y_{i} + 1$, we can choose $\bW^{3} = \ones_{q}^T$ and $\bb^{3} = -1$ so that $f_\btheta(x_i) = y_i$. This finishes the proof of Theorem~\ref{thm:htan3l} for $d_y = 1$.

\vspace*{-5pt}
\subsection{Proof for $d_y > 1$}
\vspace*{-5pt}
The proof for $d_y > 1$ is almost the same. Assume that $p \defeq d_1$ is a multiple of 2, $q \defeq d_2$ is a multiple of $2d_y$, and $pq = N d_y$.
Now partition the nodes in the 2nd layer into $d_y$ groups of size $q/d_y$. For each of the $d_y$ groups, we can do the exact same construction as done in $d_y = 1$ case, to fit each coordinate of $y_i$ perfectly. This is possible because we can share $a^{1}(x_i)$ for fitting different components of $y_i$.

%

\vspace*{-5pt}
\section{Proof of Proposition~\ref{prop:htanclsf4l}}
\vspace*{-5pt}
\label{sec:thmpf-htanclsf4l}
For the proof, we will abuse the notation slightly and let $y_i \in [d_y]$ denote the class that $x_i$ belongs to.
The idea is simple: assign distinct real numbers $\rho_1, \dots, \rho_{d_y}$ to each of the $d_y$ classes, define a new 1-dimensional regression dataset $\{ (x_i, \rho_{y_i}) \}_{i=1}^N$, and do the construction in Theorem~\ref{thm:htan3l} up to layer~2 for the new dataset. Then, we have
$\sum_{l=1}^{d_2} a_l^{2}(x_i) = \rho_{y_i} + 1$, as seen in the proof of Theorem~\ref{thm:htan3l}.

Now, at layer~3, consider the following ``gate'' activation function $\gateact$, which allows values in $(-1,+1)$ to ``pass,'' while blocking others. This can be implemented with two $\htanhact$'s or four $\relulikeact$'s:
\begin{align*}
\gateact(t) &\defeq \begin{cases}
	t+1 & -1 \leq t \leq 0,\\
	-t+1 & 0 \leq t \leq 1,\\
	0 & \text{otherwise.}
\end{cases} = \half(\htanhact(2t+1)+\htanhact(-2t+1)).
\end{align*}
For each class $j \in [d_y]$, we can choose appropriate parameters to implement a gate that allows $\rho_j$ to ``pass'' the gate, while blocking any other $\rho_{j'}$, $j' \neq j$. The output of the gate is then connected to the $j$-th output node of the network. This way, we can perfectly recover the one-hot representation for each data point.

\vspace*{-5pt}
\section{Proof of Theorem~\ref{thm:neg}}
\vspace*{-5pt}
\label{sec:thmpf-neg}
Our proof is based on the idea of counting the number of pieces of piecewise linear functions by \citet{telgarsky2015representation}. Consider any vector $u \in \reals^{d_x}$, and define the following dataset:
$x_i = i u,~~y_i = (-1)^i$, for all $i \in [N]$.

With piecewise linear activation functions, the network output $f_\btheta(x)$ is also a piecewise affine function of $x$. If we define $\bar f_\btheta(t) \defeq f_\btheta(t u)$, $\bar f_\btheta(t)$ must have at least $N-1$ linear pieces to be able to fit the given dataset $\{(x_i, y_i)\}_{i=1}^N$.
We will prove the theorem by counting the maximum number of linear pieces in $\bar f_\btheta(t)$.

We will use the following lemma, which is a slightly improved version of \citet[Lemma~2.3]{telgarsky2015representation}:
\begin{lemma}
	If $g:\reals \mapsto \reals$ and $h:\reals \mapsto \reals$ are piecewise linear with $k$ and $l$ linear pieces, respectively, then $g+h$ is piecewise linear with at most $k+l-1$ pieces, and $g \circ h$ is piecewise linear with at most $kl$ pieces.
\end{lemma}
For proof of the lemma, please refer to \citet{telgarsky2015representation}.

Consider the output of layer~1 $\bar a^{1}(t) \defeq a^{1}(t u)$, restricted for $x = tu$. For each $j \in [d_1]$, $\bar a^{1}_j(\cdot)$ has at most $p$ pieces. The input to layer~2 is a weighted sum of $\bar a^{1}_j(\cdot)$'s, so each $\bar z^{2}_k(t) \defeq z^{2}_k(tu)$ has $(p-1)d_1 + 1$ pieces, resulting in maximum $p(p-1)d_1 + p$ pieces in the corresponding output $\bar a^{2}_k(t)$. Again, the weighted sum of $d_2$ such $\bar a^{2}_k(\cdot)$'s have at most $(p(p-1)d_1 + p-1)d_2 + 1 = p(p-1)d_1 d_2 + (p-1) d_2 + 1$ pieces.

From this calculation, we can see that the output of a 2-layer network has at most $(p-1)d_1 + 1$ pieces, and a 3-layer network has $p(p-1)d_1 d_2 + (p-1) d_2 + 1$. If these number of pieces are strictly smaller than $N-1$, the network can never perfectly fit the given dataset.

\vspace*{-5pt}
\section{Proof of Proposition~\ref{thm:htanml}}
\begin{figure}[t]
	\centering
	\includegraphics[width=0.96\linewidth]{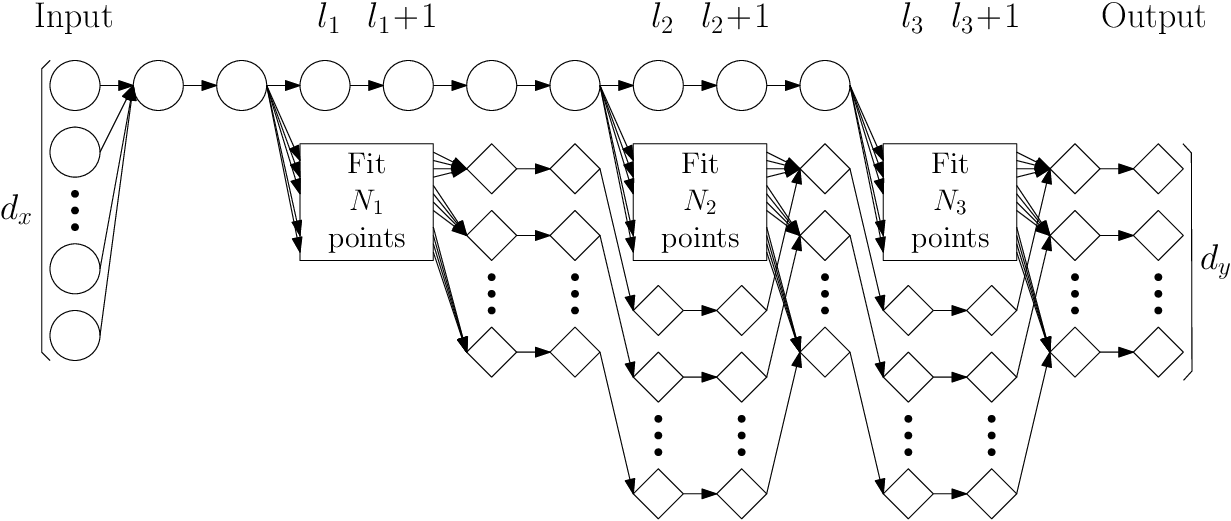}
	\vspace{18pt}
	\caption{Illustration of network parameter construction in Proposition~\ref{thm:htanml}. The circle/diamond nodes represent those carrying input/output information, respectively. The rectangular blocks are groups of nodes across two layers whose parameters are constructed from Theorem~\ref{thm:htan3l} to fit data points.}
	\vspace{-3pt}
	\label{fig:fig2}
\end{figure}

\vspace*{-5pt}
\label{sec:thmpf-htanml}
For Proposition~\ref{thm:htanml}, we will use the network from Theorem~\ref{thm:htan3l} as a building block to construct the desired parameters. The parameters we construct will result in a network illustrated in Figure~\ref{fig:fig2}. Please note that the arrows are drawn for \emph{nonzero} parameters only, and all the missing arrows just mean that the parameters are zero. We are not using a special architecture; we are still in the full connected network regime.

In the proof of Theorem~\ref{thm:htan3l}, we down-projected $x_i$'s to $u^T x_i \eqdef c_i$, and fitted $c_1, \dots, c_N$ to corresponding $y_1, \dots, y_N$. Then, what happens outside the range of the dataset? Recall from Section~\ref{sec:pf-inp-l1} that we defined $c_0 \defeq c_1 - \delta$ and $c_{N+1} \defeq c_N + \delta$ for $\delta > 0$ and constructed $\bW^{1}$ and $\bb^{1}$ using them. If we go back to the proof of Theorem~\ref{thm:htan3l}, we can check that if $u^T x \leq c_0$ or $u^T x \geq c_{N+1}$, $a_k^{2} (x) = -1$ for odd $k$'s and $+1$ for even $k$'s, resulting in
$\sum_{k=1}^q a_k^{2} (x) = 0$
for all such $x$'s. For a quick check, consider imaginary indices 0 and 17 in Figure~\ref{fig:fig1} and see which sides (left or right) of the 2nd-layer hidden nodes they will be written.

Now consider partitioning $N$ data points into $m$ subsets of cardinalities $N_1, \dots, N_m$ in the following way. We first down-project the data to get $u^T x_i$'s, and re-index data points in increasing order of $u^T x_i$'s. The first $N_1$ points go into the first subset, the next $N_2$ to the second, and so on. Then, consider constructing $m$ separate networks (by Theorem~\ref{thm:htan3l}) such that each network fits each subset, \emph{except that we let $\bb^3 = \zeros$}. As seen above, the sum of the outputs of \emph{all} these $m$ networks will be $y_i + \ones$, for all $i\in [N]$. Thus, by fitting subsets of dataset separately and summing together, we can still memorize $N$ data points. 

The rest of the proof can be explained using Figure~\ref{fig:fig2}. For simplicity, we assume that
\begin{itemize}
	\vspace*{-7pt}
	\setlength{\itemsep}{0pt}
	\item For all $j \in [m]$, $d_{l_j}-r_j$ is a multiple of 2, and $d_{l_j+1}-r_j$ is a multiple of $2d_y$,
	\item $\sum_{j=1}^{m} (d_{l_j}-r_j) (d_{l_j+1}-r_j) = N d_y$,
	\item $d_k = 1$ for all $k \in [l_1-1]$,
	\item $d_k = d_y + 1$ for all $k \in \bigcup_{j \in [m-1]}[l_j + 2:l_{j+1}-1]$,
	\item $d_k = d_y$ for all $k \in [l_m + 2:L-1]$.
	\vspace*{-7pt}
\end{itemize}
Also, let $N_j \defeq (d_{l_j}-r_j) (d_{l_j+1}-r_j)/d_y$ for $j \in [m]$.

From the input layer to layer~1, we down-project $x_i$'s using a random vector $u$, and scale $\bW^{1} \defeq u^T$ and choose $\bb^{1}$ appropriately so that $\bW^{1} x_i + \bb^{1} \in (-1,+1)$ for all $i \in [N]$. As seen in the circle nodes in Figure~\ref{fig:fig2}, this ``input information'' will be propagated up to layer $l_m-1$ to provide input data needed for fitting.

At layer $l_j - 1$, the weights and bias into the rectangular block across layers $l_j$--$(l_j+1)$ is selected in the same way as Section~\ref{sec:pf-inp-l1}. Inside each block, the subset of $N_j$ data points are fitted using the construction of Theorem~\ref{thm:htan3l}, but this time we fit to $\frac{y_i -\ones}{2}$ instead of $y_i$, in order to make sure that output information is not clipped by hard-tanh.
The output of $(l_j+1)$-th layer nodes in the block are added up and connected to diamond nodes in layer $l_j + 2$. For the $N_j$ data points in the subset, the input to the diamond nodes will be $\frac{y_i + \ones}{2}$ (instead of $y_i + \ones$), and $\zeros$ for any other data points.
As seen in Figure~\ref{fig:fig2}, this output information is propagated up to the output layer.

After fitting all $m$ subsets, the output value of diamond nodes at layer~$L-1$ is $\frac{y_i + \ones}{2}$, for all $i$. We can scale and shift this value at the output layer and get $y_i = f_\btheta(x_i)$.

\section{Proofs of Theorem~\ref{thm:genposresnet} and Corollary~\ref{cor:genposfnn}}
\subsection{Proof of Theorem~\ref{thm:genposresnet}}
\label{sec:thmpf-genpos}
The key observation used in the proof is that due to the general position assumption, if we pick any $d_x$ data points in the same class, then there always exists an affine hyperplane that contains exactly these $d_x$ points. This way, we can pick $d_x$ data points per hidden node and ``push'' them far enough to specific directions (depending on the classes), so that the last hidden layer can distinguish the classes based on the location of data points.

We use $N_k$ to denote the number of data points in class $k \in [d_y]$. Also, for $k \in [d_y]$, let $x^{\max}_{(k)}$ be the maximum value of the $k$-th component of $x_i$ over all $i \in [N]$. Also, let $\unitvec_k$ be the $k$-th standard unit vector in $\reals^{d_x}$.

Now, consider the gate activation function $\gateact$, which was also used in the proof of Proposition~\ref{prop:htanclsf4l} (Appendix~\ref{sec:thmpf-htanclsf4l}).
This activation allows values in $(-1,+1)$ to ``pass,'' while blocking others. This can be implemented with two hard-tanh ($\htanhact$) functions or four ReLU-like ($\relulikeact$) functions:
\begin{align*}
\gateact(t) &\defeq \begin{cases}
t+1 & -1 \leq t \leq 0,\\
-t+1 & 0 \leq t \leq 1,\\
0 & \text{otherwise.}
\end{cases}
= \half(\htanhact(2t+1)+\htanhact(-2t+1)).
\end{align*}
Up to layer~$L-1$, for now we will assume that the activation at the hidden nodes is $\gateact$. We will later count the actual number of hard-tanh or ReLU-like nodes required.

For class $k \in [d_y]$, we use $\lceil \frac{N_k}{d_x} \rceil$ gate hidden nodes for class $k$. Each hidden node picks and pushes $d_x$ data points in class $k$ far enough to the direction of $\unitvec_k$. Each data point is chosen only once.
Suppose that the hidden node is the $j$-th hidden node in $l$-th layer ($l \in [L-1], j \in [d_l]$). Pick $d_x$ data points in class $k$ that are not yet ``chosen,'' then there is an affine hyperplane $u^T x+c = 0$ that contains only these points.

Using the activation $\gateact$, we can make the hidden node have output $1$ for the chosen $d_x$ data points and $0$ for all remaining data points. This can be done by setting the incoming parameters
\begin{equation*}
\bU^l_{j,:} = \alpha u^T,~~\bb^l_{j} = \alpha c,
\end{equation*}
where $\alpha > 0$ is a big enough positive constant so that $|\alpha(u^T x_i + c)| > 1$ and thus $\gateact(\alpha(u^T x_i + c)) = 0$ for all unpicked data points $x_i$.
Then, choose the outgoing parameters 
\begin{equation*}
\bV^l_{:,j} = \beta \unitvec_k,~~\bc^l = \zeros
\end{equation*}
where $\beta > 0$ will be specified shortly.
Notice that since each data point is chosen only once, the $d_x$ data points were never chosen previously.
Therefore, for these $d_x$ data points, we have 
\begin{align*}
&h^{j}(x_i) = x_i,~~~~~~~~~~~\text{ for } j \in [l-1], \text{ and }\\
&h^{j}(x_i) = x_i + \beta \unitvec_k, \text{ for } j \in [l:L-1],
\end{align*}
because they will never be chosen again by other hidden nodes.
We choose big enough $\beta$ to make sure that the $k$-th component of $h^{l}(x_i)$ (i.e., $h_k^{l}(x_i)$) is bigger than $x^{\max}_{(k)} + 1$.
We also determine $\beta$ carefully so that adding $\beta \unitvec_k$ does not break the general position assumption.
The values of $\beta$ that breaks the general position lie in a set of measure zero, so we can sample $\beta$ from some suitable continuous random distribution to avoid this.

After doing this to all data points, $h^{L-1}(x_i)$ satisfies the following property:
For $x_i$'s that are in class $k$, $h_k^{L-1}(x_i) \geq x^{\max}_{(k)} + 1$,
and for $x_i$'s that are not in class $k$, $h_k^{L-1}(x_i) \leq x^{\max}_{(k)}$.

At layer $L$, by assumption we have $d_L \geq d_y$ in case of hard-tanh ResNet.
We assume $d_L = d_y$ for simplicity, and choose
\begin{align*}
&\bU^L = \begin{bmatrix}
2\cdot I_{d_y \times d_y} & 
\zeros_{d_y \times (d_x-d_y)}
\end{bmatrix},~~
\bb^L = \begin{bmatrix}
-2 x^{\max}_{(1)} - 1\\
-2 x^{\max}_{(2)} - 1\\
\vdots\\
-2 x^{\max}_{(d_y)} - 1
\end{bmatrix},
\end{align*}
then by clipping of hard-tanh, for $x_i$ in class $k$, the $k$-th component of $\sigma (\bU^L h^{L-1}(x_i) + \bb^{L})$ is $+1$ and all the other components are $-1$.
Now, by choosing
\begin{align*}
&\bV^L = \frac{1}{2} \cdot I_{d_y \times d_y}
,~~
\bc^L = \frac{1}{2} \ones_{d_y},
\end{align*}
we can recover the one-hot representation: $g_\btheta(x_i) = y_i$, for all $i \in [N]$.
For ReLU-like ResNets, we can do the same job by using $d_L = 2d_y$.

Finally, let us count the number of hidden nodes used, for layers up to $L-1$. 
Recall that we use $\lceil \frac{N_k}{d_x} \rceil$ \textbf{gate} activation nodes for class $k$.
Note that the total number of gate activations used is bounded above by
\begin{equation*}
\sum_{k=1}^{d_y} \left \lceil \frac{N_k}{d_x} \right \rceil 
\leq \sum_{k=1}^{d_y} \left ( \frac{N_k}{d_x} + 1 \right )
= \frac{N}{d_x} + d_y,
\end{equation*}
and each gate activation can be constructed with two hard-tanh nodes or four ReLU-like nodes.
Therefore, $\sum_{l=1}^{L-1} d_l \geq \frac{2N}{d_x}+2d_y$ and $d_L \geq d_y$ is the sufficient condition for
a hard-tanh ResNet to realize the above construction, and ReLU-like ResNets require twice as many hidden nodes.

\subsection{Proof of Corollary~\ref{cor:genposfnn}}
\label{sec:corpf-genposfnn}
The main idea of the proof is exactly the same. We use $\lceil \frac{N_k}{d_x} \rceil$ gate activation nodes for class $k$,
and choose $d_x$ data points in the same class per each hidden node. When the hidden node is the $j$-th node in the hidden layer and the chosen points are from class $k$, we choose
\begin{equation*}
\bW^2_{:,j} = \unitvec_k,~\bb^2 = \zeros.
\end{equation*}
This way, one can easily recover the one-hot representation and achieve $f_\btheta(x_i) = y_i$.

\section{Proof of Theorem~\ref{thm:memorizer}}
\label{sec:thmpf-memorizer}
The outline of the proof is as follows.
Recall that we write $\btheta^{(t)}$ as $\btheta^* + \bxi^{(t)}$. By the chain rule, we have
\begin{equation*}
\nabla_{\btheta} \erisk(\btheta^*+\bxi^{(t)}) 
= \frac{1}{N} \sum_{i=1}^N 
\ell'(f_{\btheta^*+\bxi^{(t)}}(x_i);y_i)
\nabla_\btheta f_{\btheta^*+\bxi^{(t)}}(x_i).
\end{equation*}
If $\bxi^{(t)}$ is small enough, the terms $\ell'(f_{\btheta^*+\bxi^{(t)}}(x_i);y_i)$ and $\nabla_\btheta f_{\btheta^*+\bxi^{(t)}}(x_i)$ can be expressed in terms of perturbation on $\ell'(f_{\btheta^*}(x_i);y_i)$ and $\nabla_\btheta f_{\btheta^*}(x_i)$, respectively (Lemma~\ref{lem:GDtaylor}).
We then use the lemma and prove each statement of the theorem.

We first begin by introducing more definitions and symbols required for the proof.
As mentioned in the main text, we'll abuse the notation $\btheta$ to mean the concatenation of vectorizations of all the parameters $(\bW^l, \bb^l)_{l=1}^L$.
To simplify the notation, we define $\ell_i (\btheta) \defeq \ell(f_\btheta(x_i);y_i)$. Same thing applies for derivatives of $\ell$: $\ell'_i (\btheta) \defeq \ell'(f_\btheta(x_i);y_i)$, and so on.

Now, for each data point $i \in [N]$ and each layer $l \in [L-1]$, define the following diagonal matrix:
\begin{equation*}
J_{\btheta}^l (x_i) \defeq \diag(
\begin{bmatrix}
\sigma'(z_1^l(x_i)) &
\cdots&
\sigma'(z_{d_l}^l(x_i))
\end{bmatrix}
) 
\in \reals^{d_l \times d_l},
\end{equation*}
where $\sigma'$ is the derivative of the activation function $\sigma$, wherever it exists.

Now consider a memorizing global minimum $\btheta^*$. 
As done in the main text, we will express any other point $\btheta$ as
$\btheta = \btheta^* + \bxi$,
where $\bxi$ is the vectorized version of perturbations. 
By assumption, $\erisk(\cdot)$ is differentiable at $\btheta^*$;
this means that $J_{\btheta^*}^l(x_i)$ are well-defined at $\btheta^*$ for all data points and layers $l \in [L-1]$.
Moreover, since $\sigma$ is piecewise linear, there exists a small enough positive constant $\rho_c$ such that
for any $\bxi$ satisfying $\norms{\bxi} \leq \rho_c$, the slopes of activation functions stay constant, 
i.e., $J_{\btheta^*+\bxi}^l(x_i) = J_{\btheta^*}^l(x_i)$ for all $i \in [N]$ and $l \in [L-1]$.


Now, as in the main text, define vectors $\nu_i \defeq \nabla_\btheta f_{\btheta^*}(x_i)$ for all $i \in [N]$.
We can then express $\bxi$ as the sum of two orthogonal components $\bxi_{\parallel}$ and $\bxi_{\perp}$,
where $\bxi_{\parallel} \in \spann(\{\nu_i\}_{i=1}^N)$ and $\bxi_{\perp} \in \spann(\{\nu_i\}_{i=1}^N)^{\perp}$.
We also define $P_\nu$ to be the projection matrix onto $\spann(\{\nu_i\}_{i=1}^N)$; note that $\bxi_{\parallel} = P_\nu \bxi$.

Using the fact that perturbations are small, we can calculate the deviation of network output $f_{\btheta^*+\bxi} (x_i)$ from $f_{\btheta^*} (x_i)$, and use Taylor expansion of $\ell$ and $\ell'$ to show the following lemma, whose proof is deferred to Appendix~\ref{sec:lempf-GDtaylor}.

\begin{lemma}
	\label{lem:GDtaylor}
	For any given memorizing global minimum $\btheta^*$ of $\erisk(\cdot)$,
	there exist positive constants $\rho_s$ \textup{($\leq \rho_c$)}, $C_1$, $C_2$, $C_3$, $C_4$, and $C_5$ such that,
	if $\norms{\bxi} \leq \rho_s$, the following holds for all $i \in [N]$:
	\begin{align*}
	&\ell_i(\btheta^*+\bxi) - \ell_i(\btheta^*) \leq C_1 (C_2\norms{\bxi_{\parallel}} + C_3 \norms{\bxi}^2)^2,\\
	&\ell'_i(\btheta^*+\bxi) = \ell''_i(\btheta^*) \nu_i^T \bxi_{\parallel} + R_{i}(\bxi),\\
	&\nabla_\btheta f_{\btheta^*+\bxi}(x_i) = \nu_i + \mu_i(\bxi),
	\end{align*}
	where the remainder/perturbation terms satisfy
	\begin{align*}
	|R_i(\bxi)| \leq C_4 \norms{\bxi}^2, \text{ and }
	\norms{\mu_i(\bxi)} \leq C_5 \norms{\bxi}.
	\end{align*}
\end{lemma}
Besides the constants defined in Lemma~\ref{lem:GDtaylor}, define
\begin{equation*}
C_6 \defeq \max_{i \in [N]} \ell''_i(\btheta^*) \norms {\nu_i}.
\end{equation*} 
Also, it will be shown in the proof of Lemma~\ref{lem:GDtaylor} that $C_2 \defeq \max_{i \in [N]} \norms {\nu_i}$.
Given Lemma~\ref{lem:GDtaylor}, we are now ready to prove Theorem~\ref{thm:memorizer}.

Let us first consider the case where all $\nu_i$'s are zero vectors, so $\spann(\{\nu_i\}_{i=1}^N) = \{\zeros{}\}$.
For such a pathological case, $\bxi_{\parallel}^{(0)} = \zeros{}$, so the condition $\norms{\bxi_{\parallel}^{(t)}} \geq \tau \norms{\bxi^{(t)}}^2$ is 
violated at $t^* = 0$ for any positive $\tau$. By Lemma~\ref{lem:GDtaylor},
\begin{equation*}
\ell_i(\btheta^*+\bxi^{(0)}) - \ell_i(\btheta^*) \leq C_1 C_3^2 \norms{\bxi^{(0)}}^4,
\end{equation*}
as desired; for this case, Theorem~\ref{thm:memorizer} is proved with $\rho \defeq \rho_s$, $C \defeq C_1 C_3^2$.

For the remaining case where $\spann(\{\nu_i\}_{i=1}^N) \neq \{\zeros{}\}$, let
$H \defeq \sum_{i=1}^N \ell''_i (\btheta^*) \nu_i \nu_i^T$, and define $\lambda_{\min}$ and $\lambda_{\max}$ to be the smallest and largest strictly positive eigenvalues of $H$, respectively. We will show that Theorem~\ref{thm:memorizer} holds with the following constant values:
\begin{align*}
\tau &\defeq \frac{16C_2 C_4 N}{\lambda_{\min}},\\
\rho &\defeq \frac{1}{2} \min \left \{\rho_s, \frac{\lambda_{\min} C_2}{16 C_2 C_5 C_6 N + \lambda_{\min} C_5 } \right \}.\\
\gamma &\defeq \min \left \{ \frac{8B \log 2}{\lambda_{\min}}, \frac{\lambda_{\min} B}{2 \lambda_{\max}^2 \epoch^2} \right \},\\
\lambda &\defeq \frac{\lambda_{\min}}{4B},\\
C &\defeq 16 C_1(C_2 \tau + C_3)^2.
\end{align*}
Firstly, as we saw in the previous case, if $\norms{\bxi_{\parallel}^{(t)}} \geq \tau \norms{\bxi^{(t)}}^2$ is 
violated at $t^* = 0$, we immediately have
\begin{equation*}
\ell_i(\btheta^*+\bxi^{(0)}) - \ell_i(\btheta^*) \leq C_1 (C_2 \tau+C_3)^2 \norms{\bxi^{(0)}}^4 \leq C \norms{\bxi^{(0)}}^4.
\end{equation*}

Now suppose $\norms{\bxi_{\parallel}^{(t)}} \geq \tau \norms{\bxi}^2$ is satisfied up to some iterations, so $t^* > 0$. We will first prove that as long as $(k+1)\epoch \leq t^*$, we have
\begin{equation*}
\norms {\bxi_{\parallel}^{(k\epoch+\epoch)}} \leq (1-\eta \lambda) \norms {\bxi_{\parallel}^{(k\epoch)}}.
\end{equation*}
To simplify the notation, we will prove this for $k = 0$; as long as $(k+1)\epoch \leq t^*$, the proof extends to other values of $k$.

Using Lemma~\ref{lem:GDtaylor}, we can write the gradient estimate $g^{(t)}$ at $\btheta^{(t)} = \btheta^* + \bxi^{(t)}$ as:
\begin{align*}
&g^{(t)}
= \frac{1}{B} \sum_{i \in B^{(t)}} 
\ell'_i(\btheta^*+\bxi^{(t)}) \nabla_\btheta f_{\btheta^*+\bxi^{(t)}}(x_i)\\
=& \frac{1}{B} \sum_{i \in B^{(t)}} 
\left ( 
\ell''_i(\btheta^*) \nu_i^T \bxi_{\parallel}^{(t)} + R_{i}(\bxi^{(t)})
\right) 
\left ( 
\nu_i + \mu_i(\bxi^{(t)})
\right)\\
=& 
\Bigg ( 
\frac{1}{B} \sum_{i \in B^{(t)}} 
\ell''_i(\btheta^*) \nu_i \nu_i^T
\Bigg ) \bxi_{\parallel}^{(t)} +
\underbrace{
\frac{1}{B} \sum_{i \in B^{(t)}} 
	\left (
	\ell''_i(\btheta^*) \nu_i^T \bxi_{\parallel}^{(t)} \mu_i(\bxi^{(t)})+
	R_{i}(\bxi^{(t)}) ( \nu_i + \mu_i(\bxi^{(t)}))
	\right) }_{\eqdef \bzeta^{(t)}}.
\end{align*}

After the SGD update $\btheta^{(t+1)} \leftarrow \btheta^{(t)} - \eta g^{(t)}$,
\begin{align*}
\btheta^*+\bxi_{\parallel}^{(t+1)} + \bxi_{\perp}^{(t+1)}
&=\btheta^*+\bxi_{\parallel}^{(t)} + \bxi_{\perp}^{(t)} - \eta g^{(t)} \\
&=\btheta^*+
\Bigg( 
I - \frac{\eta}{B} \sum_{i \in B^{(t)}}
\ell''_i(\btheta^*) \nu_i \nu_i^T
\Bigg) \bxi_{\parallel}^{(t)}
+ \bxi_{\perp}^{(t)} 
- \eta \bzeta^{(t)}.
\end{align*}
Since $\eta < \gamma \leq \frac{B}{\lambda_{\max}}$, $I - \frac{\eta}{B} \sum_{i \in B^{(t)}} \ell''_i(\btheta^*) \nu_i \nu_i^T$ is a positive semi-definite matrix with spectral norm at most $1$. Using the projection matrix $P_\nu$, we can write
\begin{align}
\label{eq:update1}
\bxi_{\parallel}^{(t+1)} &= 
\Bigg( 
I - \frac{\eta}{B} \sum_{i \in B^{(t)}}
\ell''_i(\btheta^*) \nu_i \nu_i^T
\Bigg) \bxi_{\parallel}^{(t)}
- \eta P_\nu \bzeta^{(t)},\\
\label{eq:update2}
\bxi_{\perp}^{(t+1)} &=
\bxi_{\perp}^{(t)} 
- \eta (I-P_\nu) \bzeta^{(t)}.
\end{align}

Now, by Lemma~\ref{lem:GDtaylor},
\begin{align*}
\norms{\bzeta^{(t)}} 
\leq& \frac{1}{B} \sum_{i \in B^{(t)}} 
\left (
\norms{\ell''_i(\btheta^*) \nu_i^T \bxi_{\parallel}^{(t)} \mu_i(\bxi^{(t)})} + \norms{R_{i}(\bxi^{(t)}) \nu_i} + \norms{R_{i}(\bxi^{(t)}) \mu_i(\bxi^{(t)})}
\right )\\
\leq& C_5 C_6 \norms{\bxi^{(t)}}  \norms{\bxi_{\parallel}^{(t)}} + C_2 C_4 \norms{\bxi^{(t)}}^2 + C_4 C_5 \norms{\bxi^{(t)}}^3.
\end{align*}
Under the condition that $\norms{\bxi_{\parallel}^{(t)}} \geq \tau \norms{\bxi^{(t)}}^2$, where $\tau \defeq \frac{16 C_2 C_4 N}{\lambda_{\min}}$, 
and also that $\norms{\bxi^{(t)}} \leq \rho \leq \frac{\lambda_{\min} C_2}{16 C_2 C_5 C_6 N + \lambda_{\min} C_5}$,
\begin{align*}
\norms{\bzeta^{(t)}}
&\leq \frac{C_2 C_4}{\tau} \norms{\bxi_{\parallel}^{(t)}} + \left (C_5 C_6 + \frac{C_4 C_5}{\tau} \right ) \norms{\bxi^{(t)}} \norms{\bxi_{\parallel}^{(t)}}\\
&\leq \left (\frac{\lambda_{\min}}{16N} + \left (C_5 C_6 + \frac{\lambda_{\min} C_5}{16C_2 N} \right ) \norms{\bxi^{(t)}} \right )\norms{\bxi_{\parallel}^{(t)}}
\leq \frac{\lambda_{\min}}{8N} \norms{\bxi_{\parallel}^{(t)}}.
\end{align*}
From this, we can see that 
\begin{equation*}
\norms{\bxi_{\parallel}^{(t+1)}}
\leq
\norms{\bxi_{\parallel}^{(t)}}
+ \eta \norms{\bzeta^{(t)}} 
\leq \left ( 1 + \frac{\eta \lambda_{\min}}{8N} \right ) \norms{\bxi_{\parallel}^{(t)}}.
\end{equation*}
Noting that $\eta < \gamma \leq \frac{8 B \log 2}{\lambda_{\min}}$, 
\begin{equation*}
\left ( 1 + \frac{\eta \lambda_{\min}}{8N} \right )^\epoch \leq
\left ( 1 + \frac{\log 2}{\epoch} \right )^\epoch \leq 2,
\end{equation*}
so for $1 \leq t \leq \epoch$,
\begin{equation*}
\norms{\bzeta^{(t)}}
\leq
\frac{\lambda_{\min}}{8N} \norms{\bxi_{\parallel}^{(t)}}
\leq
\frac{\lambda_{\min}}{8N}
\left ( 1 + \frac{\log 2}{\epoch} \right )^t \norms{\bxi_{\parallel}^{(0)}}
\leq 
\frac{\lambda_{\min}}{4N} \norms{\bxi_{\parallel}^{(0)}}.
\end{equation*}

Now, repeating the update rule~\eqref{eq:update1} from $t = 0$ to $\epoch-1$, we get
\begin{equation}
\label{eq:updateE1}
\bxi_{\parallel}^{(\epoch)} = 
\prod_{k = \epoch-1}^0
\Big( 
I - \frac{\eta}{B} H_k
\Big) 
\bxi_{\parallel}^{(0)}
- \eta
\sum_{t=0}^{\epoch-1} 
\prod_{k = \epoch-1}^{t+1}
\Big( 
I - \frac{\eta}{B} H_k
\Big)
P_\nu \bzeta^{(t)},
\end{equation}
where $H_k \defeq \sum_{i \in B^{(k)}}
\ell''_i(\btheta^*) \nu_i \nu_i^T$.
We are going to bound the norm of each term.
For the second term, we have
\begin{equation}
\label{eq:2nd}
\norm{
\sum_{t=0}^{\epoch-1} 
\prod_{k = \epoch-1}^{t+1}
\Big( 
I - \frac{\eta}{B} H_k
\Big)
P_\nu \bzeta^{(t)}
}
\leq \sum_{t=0}^{\epoch-1} \norms{\bzeta^{(t)}}
\leq \frac{\lambda_{\min}E}{4N} \norms{\bxi_{\parallel}^{(0)}}
= \frac{\lambda_{\min}}{4B} \norms{\bxi_{\parallel}^{(0)}}.
\end{equation}
The first term is a bit tricker. Note first that
\begin{align*}
\prod_{k = \epoch-1}^0
\Big( 
I - \frac{\eta}{B} H_k
\Big) 
= I - \frac{\eta}{B} \sum_{k=0}^{\epoch-1} H_k 
+ \frac{\eta^2}{B^2} \sum_{\substack{j,k \in [0,\epoch-1] \\ j<k }} H_k H_j 
- \frac{\eta^3}{B^3} \sum_{\substack{i,j,k \in [0,\epoch-1] \\ i<j<k }} H_k H_j H_i + \cdots.
\end{align*}
Recall the definition $H = \sum_{i=1}^N \ell''_i(\btheta^*) \nu_i \nu_i^T = \sum_{k=0}^{\epoch-1} H_k$, and that $\lambda_{\min}$ and $\lambda_{\max}$ are the minimum and maximum eigenvalues of $H$. Since $H_k$'s are positive semi-definite and $H$ is the sum of $H_k$'s, the maximum eigenvalue of $H_k$ is at most $\lambda_{\max}$. Using this,
\begin{align*}
\norm{
\prod_{k = \epoch-1}^0
\Big( 
I - \frac{\eta}{B} H_k
\Big) 
\bxi_{\parallel}^{(0)}
}
\leq 
\left (
1-\frac{\eta \lambda_{\min}}{B}
+ \sum_{k=2}^E \choose{\epoch}{k} \Big ( \frac{\eta \lambda_{\max}}{B} \Big )^k
\right )
\norms{\bxi_{\parallel}^{(0)}}.
\end{align*}
First note that for $k \in [2, E-1]$, $\choose{E}{k+1} \frac{2}{E} \leq \choose {E}{k}$, because
\begin{align*}
\frac{2}{E} \leq \frac{k+1}{E-k} = 
\frac{(k+1)! (E-k-1)!}{k! (E-k)!} =
\frac{\choose {E}{k}}{\choose {E}{k+1}}.
\end{align*}
Since $\eta < \gamma \leq \frac{\lambda_{\min} B}{2 \lambda_{\max}^2 \epoch^2} \leq \frac{B}{\lambda_{\max}\epoch}$, for $k \in [2, E-1]$ we have
\begin{align*}
\choose{E}{k+1} \Big ( \frac{\eta \lambda_{\max}}{B} \Big )^{k+1}
\leq \choose{E}{k+1} \frac{1}{E} \Big ( \frac{\eta \lambda_{\max}}{B} \Big )^{k}
\leq \frac{1}{2} \choose{E}{k} \Big ( \frac{\eta \lambda_{\max}}{B} \Big )^{k},
\end{align*}
which implies that
\begin{align*}
\sum_{k=2}^E \choose{\epoch}{k} \Big ( \frac{\eta \lambda_{\max}}{B} \Big )^k
\leq 2 \choose{\epoch}{2} \Big ( \frac{\eta \lambda_{\max}}{B} \Big )^2
\leq \frac{\eta^2 E^2 \lambda_{\max}^2}{B^2} 
\leq \frac{\eta \lambda_{\min}}{2B}.
\end{align*}
Therefore, we have 
\begin{equation*}
\norm{
	\prod_{k = \epoch-1}^0
	\Big( 
	I - \frac{\eta}{B} H_k
	\Big) 
	\bxi_{\parallel}^{(0)}
}
\leq 
\left (
1-\frac{\eta \lambda_{\min}}{2B}
\right )
\norms{\bxi_{\parallel}^{(0)}}.
\end{equation*}
Together with the bound on the second term~\eqref{eq:2nd}, this shows that
\begin{equation*}
\norms{\bxi_{\parallel}^{(\epoch)}}
\leq 
\left (
1-\frac{\eta \lambda_{\min}}{4B}
\right )
\norms{\bxi_{\parallel}^{(0)}}
= 
\left (
1-\eta \lambda
\right )
\norms{\bxi_{\parallel}^{(0)}},
\end{equation*}
which we wanted to prove.

We now have to prove that
\begin{equation*}
\norms{\bxi^{(\epoch)}}
\leq
\norms{\bxi^{(0)}}
+ \eta \lambda \norms{\bxi_{\parallel}^{(0)}}.
\end{equation*}
Now, repeating the update rule~\eqref{eq:update2} from $t = 0$ to $\epoch-1$, we get
\begin{equation}
\label{eq:updateE2}
\bxi_{\perp}^{(\epoch)} = 
\bxi_{\perp}^{(0)}
- \eta
\sum_{t=0}^{\epoch-1} 
(I-P_\nu) \bzeta^{(t)}.
\end{equation}
Thus, by combining equations~\eqref{eq:updateE1} and \eqref{eq:updateE2},
\begin{align*}
&\norms{\bxi^{(\epoch)}}
= \norms{\bxi_{\parallel}^{(\epoch)}+\bxi_{\perp}^{(\epoch)}}\\
\leq& \norm{
\prod_{k = \epoch-1}^0
\Big( 
I - \frac{\eta}{B} H_k
\Big) 
\bxi_{\parallel}^{(0)} +\bxi_{\perp}^{(0)} 
}
+ \eta \sum_{t=0}^{E-1}
\norm{
\prod_{k = \epoch-1}^{t+1}
\Big( 
I - \frac{\eta}{B} H_k
\Big)
P_\nu \bzeta^{(t)}
+
(I-P_\nu) \bzeta^{(t)}
}\\
\leq& \norms{\bxi^{(0)}}
+ \eta \sum_{t=0}^{E-1}
\norms{\bzeta^{(t)}} 
\leq 
\norms{\bxi^{(0)}}
+ \eta \frac{\lambda_{\min}}{4B} \norms{\bxi_{\parallel}^{(0)}}
= \norms{\bxi^{(0)}} + \eta \lambda \norms{\bxi_{\parallel}^{(0)}}.
\end{align*}

It now remains to prove that $\norms {\bxi^{(t^*)}} \leq 2\norms {\bxi^{(0)}} \leq 2\rho$ at the first iteration $t^*$ that $\norms{\bxi_{\parallel}^{(t)}} \geq \tau \norms{\bxi^{(t)}}^2$ is violated. Let $k^*$ be the maximum $k$ such that $k\epoch \leq t^*$.

From what we have shown so far,
\begin{equation*}
\norms{\bxi^{(k^*\epoch)}} 
\leq \norms{\bxi^{(0)}} + \eta \lambda \sum_{k=0}^{k^*-1} \norms{\bxi_{\parallel}^{(k\epoch)}}.
\end{equation*}
Also, for $t$ in $k^* \epoch \leq t < t^*$ the condition $\norms{\bxi_{\parallel}^{(t)}} \geq \tau \norms{\bxi^{(t)}}^2$ is satisfied, so by the same argument we have $\norms{\bzeta^{(t)}} \leq \frac{\lambda_{\min}}{4N} \norms{\bxi_{\parallel}^{(k^*\epoch)}}$ for $t \in [k^*E , t^*-1]$.
Finally, by modifying equations~\eqref{eq:updateE1} and \eqref{eq:updateE2} a bit, we get
\begin{align*}
&\norms{\bxi^{(t^*)}}
= \norms{\bxi_{\parallel}^{(t^*)}+\bxi_{\perp}^{(t^*)}}\\
\leq& \norm{
	\prod_{k = t^*-1}^{k^*E}
	\Big( 
	I - \frac{\eta}{B} H_k
	\Big) 
	\bxi_{\parallel}^{(k^* E)} +\bxi_{\perp}^{(k^* E)} 
}
+ \eta \sum_{t=k^*E}^{t^*-1}
\norm{
	\prod_{k = t^*-1}^{t+1}
	\Big( 
	I - \frac{\eta}{B} H_k
	\Big)
	P_\nu \bzeta^{(t)}
	+
	(I-P_\nu) \bzeta^{(t)}
}\\
\leq& \norms{\bxi^{(k^* E)}}
+ \eta \sum_{t=k^* E}^{t^*-1}
\norms{\bzeta^{(t)}} 
\leq 
\norms{\bxi^{(k^* E)}} + \eta \frac{\lambda_{\min}}{4B} \norms{\bxi_{\parallel}^{(k^* E)}}
\leq \norms{\bxi^{(0)}} + \eta \lambda \sum_{k=0}^{k^*} \norms{\bxi_{\parallel}^{(k\epoch)}}.
\end{align*}
Finally, from $\norms {\bxi_{\parallel}^{(k\epoch+\epoch)}} \leq (1-\eta \lambda) \norms {\bxi_{\parallel}^{(k\epoch)}}$,
\begin{align*}
\norms{\bxi^{(t^*)}}
\leq \norms{\bxi^{(0)}} + \eta \lambda \sum_{k=0}^{k^*} (1-\eta \lambda)^k \norms{\bxi_{\parallel}^{(0)}}
\leq \norms{\bxi^{(0)}} + \norms{\bxi_{\parallel}^{(0)}}
\leq 2\norms{\bxi^{(0)}}.
\end{align*}

\section{Proof of Lemma~\ref{lem:st3matrix}}
\label{sec:lemst3matrix}

Recall that $i_{k,j} \in [jq-q+1, jq]$. Consider any $l < j$. Then, $i_{k,j} > lq$, so by \eqref{eq:st2clipoddmore} and \eqref{eq:st2clipevenmore}, we have $a_{l}^{1}(x_{i_{k,j}}) = (-1)^{l-1}$. Similarly, if we consider $l > j$, then $i_{k,j} \leq lq-q$, so it follows from \eqref{eq:st2clipoddless} and \eqref{eq:st2clipevenless} that $a_{l}^{1}(x_{i_{k,j}}) = (-1)^{l}$. 
This means that the entries (indexed by $(j,l)$) of $M_k$ below the diagonal are filled with $(-1)^{l-1}$, and entries above the diagonal are filled with $(-1)^l$. Thus, the matrix $M_k$ has the form
\begin{align*}
M_k = 
\begin{bmatrix}
a_1^{1}(x_{i_{k,1}}) & 1 & -1 & \cdots & -1 & 1 & 1 \\
1 & a_2^{1}(x_{i_{k,2}}) & -1 & \cdots & -1 & 1 & 1 \\
1 & -1 & a_3^{1}(x_{i_{k,3}}) & \cdots & -1 & 1 & 1 \\
\vdots & \vdots & \vdots & \ddots & \vdots& \vdots& \vdots\\
1 & -1 & 1 & \cdots & a_{p-1}^{1}(x_{i_{k,p-1}}) & 1 & 1\\
1 & -1 & 1 & \cdots & 1 & a_p^{1}(x_{i_{k,p}}) & 1
\end{bmatrix}.
\end{align*}

To prove the first statement of Lemma~\ref{lem:st3matrix}, consider adding the last column to every even $l$-th column and subtracting it from every odd $l$-th column. Then, this results in a matrix
\begin{align*}
\begin{bmatrix}
a_1^{1}(x_{i_{k,1}})-1 & 2 & -2 & \cdots & -2 & 2 & 1 \\
0 & a_2^{1}(x_{i_{k,2}})+1 & -2 & \cdots & -2 & 2 & 1 \\
0 & 0 & a_3^{1}(x_{i_{k,3}})-1 & \cdots & -2 & 2 & 1 \\
\vdots & \vdots & \vdots & \ddots & \vdots& \vdots& \vdots\\
0 & 0 & 0 & \cdots & a_{p-1}^{1}(x_{i_{k,p-1}})-1 & 2 & 1\\
0 & 0 & 0 & \cdots & 0& a_p^{1}(x_{i_{k,p}})+1 & 1
\end{bmatrix},
\end{align*}
whose columns space is the same as $M_k$. It follows from $a_j^{1}(x_{i_{k,j}}) \in (-1,+1)$ that $M_k$ has full column rank. This also implies that $\dim(\nulsp(M_k)) = 1$.

For the second statement, consider subtracting ($j+1$)-th row from $j$-th row, for $j \in [p-1]$. This results in
\begin{align*}
\tilde M_k \defeq
\begin{bmatrix}
a_1^{1}(x_{i_{k,1}})-1 & 1-a_2^{1}(x_{i_{k,2}}) & 0 & \cdots & 0 & 0 & 0 \\
0 & a_2^{1}(x_{i_{k,2}})+1 & -a_3^{1}(x_{i_{k,3}})-1 & \cdots & 0 & 0 & 0 \\
0 & 0 & a_3^{1}(x_{i_{k,3}})-1 & \cdots & 0 & 0 & 0 \\
\vdots & \vdots & \vdots & \ddots & \vdots& \vdots& \vdots\\
0 & 0 & 0 & \cdots & a_{p-1}^{1}(x_{i_{k,p-1}})-1 & 1-a_p^{1}(x_{i_{k,p}}) & 0\\
1 & -1 & 1 & \cdots & 1 & a_p^{1}(x_{i_{k,p}}) & 1
\end{bmatrix},
\end{align*}
which has the same null space as $M_k$. Consider a nonzero vector $\nu \in \nulsp(\tilde M_k)$, i.e., $\tilde M_k \nu = \zeros$. 
Let $\nu_l$ denote the $l$-th component of $\nu$. One can see that $\nu_1, \dots, \nu_p$ are not all zero, because if $\nu_{p+1}$ is the only nonzero component, $\tilde M_k \nu = (0,0,\dots,0,\nu_{p+1})^T \neq \zeros$.

Assume without loss of generality that $\nu_1$ is strictly positive.
Note that $a_1^{1}(x_{i_{k,1}})-1$ and $1-a_2^{1}(x_{i_{k,2}})$ are both nonzero and the signs of $a_1^{1}(x_{i_{k,1}})-1$ and $1-a_2^{1}(x_{i_{k,2}})$ are opposite. Then if follows from $(a_1^{1}(x_{i_{k,1}})-1) \nu_1 + (1-a_2^{1}(x_{i_{k,2}})) \nu_2 = 0$ that $\nu_2$ is also strictly positive.
Similarly, $a_2^{1}(x_{i_{k,2}})+1$ and $-a_3^{1}(x_{i_{k,3}})-1$ are both nonzero and have opposite signs, so $\nu_3 > 0$. Proceeding this way up to $\nu_p$, we can see that all $\nu_l$, $l \in [p]$, are strictly positive.

\section{Proof of Lemma~\ref{lem:GDtaylor}}
\label{sec:lempf-GDtaylor}
We begin by introducing more definitions.
For a matrix $A \in \reals^{m \times n}$, let $\vect(A) \in \reals^{mn}$ be its vectorization, i.e., columns of $A$ concatenated as a long vector.
Given matrices $A$ and $B$, let $A \otimes B$ denote their Kronecker product.
Throughout the proof, we use $\btheta$ and $\bxi$ to denote the concatenation of vectorizations of all the parameters $(\bW^l, \bb^l)_{l=1}^L$ and perturbations $(\bDelta^l, \bdelta^l)_{l=1}^L$:
\begin{equation}
\label{eqn:thetaxivec}
\btheta \defeq
\begin{bmatrix}
\vect(\bW^L)\\
\bb^L\\
\vect(\bW^{L-1})\\
\bb^{L-1}\\
\vdots\\
\vect(\bW^{1})\\
\bb^{1}
\end{bmatrix},~~~~~
\bxi \defeq
\begin{bmatrix}
\vect(\bDelta^L)\\
\bdelta^L\\
\vect(\bDelta^{L-1})\\
\bdelta^{L-1}\\
\vdots\\
\vect(\bDelta^{1})\\
\bdelta^{1}
\end{bmatrix}.
\end{equation}
In Section~\ref{sec:prelim}, we defined $a^l(x_i)$ to denote output of the $l$-th hidden layer when the network input is $x_i$. In order to make the dependence of parameters more explicit, we will instead write $a_{\btheta}^l(x_i)$ in this section.
Also, for $l \in [L-1]$, define
\begin{equation}
\label{eqn:Dmatrix}
D_{\btheta}^l (x_i)
\defeq W^L J_{\btheta}^{L-1} (x_i) W^{L-1} \cdots W^{l+1} J_{\btheta}^{l}(x_i) \in \reals^{1 \times d_l},
\end{equation}
and for convenience in notation, let $D_{\btheta}^L (x_i) \defeq 1$.
It can be seen from standard matrix calculus that
\begin{equation}
\label{eqn:gradfxilayerl}
\begin{bmatrix}
\nabla_{\bW^l} f_{\btheta}(x_i)
& 
\nabla_{\bb^l} f_{\btheta}(x_i)
\end{bmatrix}
= D_{\btheta}^l (x_i)^T  
\begin{bmatrix}
a_{\btheta}^{l-1}(x_i)^T 
&
1
\end{bmatrix},
\end{equation}
for all $l \in [L]$.
Vectorizing and concatenating these partial derivatives results in
\begin{equation}
\label{eqn:gradfxi}
\nabla_\btheta f_\btheta(x_i)=
\begin{bmatrix}
a_{\btheta}^{L-1}(x_i)\\
1\\
\begin{bmatrix}
a_{\btheta}^{L-2}(x_i)\\
1
\end{bmatrix}
\otimes
D_{\btheta}^{L-1}(x_i)^T\\
\vdots\\
\begin{bmatrix}
x_i\\
1
\end{bmatrix}
\otimes
D_{\btheta}^{1}(x_i)^T\\
\end{bmatrix}.
\end{equation}

In order to prove the lemma, we first have to quantify how perturbations on the global minimum affect outputs of the hidden layers and the network.
Let $\btheta^* \defeq (\bW_*^l, \bb_*^l)_{l=1}^L$ be the memorizing global minimum, and let $(\bDelta^l, \bdelta^l)_{l=1}^L$ be perturbations on parameters, whose vectorization $\bxi$ satisfies $\norms{\bxi} \leq \rho_c$. Then, for all $l \in [L-1]$, define $\tilde a^l_{\btheta^*+\bxi}(\cdot)$ to denote the amount of perturbation in the $l$-th hidden layer output:
\begin{equation*}
\tilde a^l_{\btheta^*+\bxi} (x_i) \defeq a^l_{\btheta^*+\bxi} (x_i) - a^l_{\btheta^*} (x_i).
\end{equation*}
It is easy to check that
\begin{align*}
&\tilde a^1_{\btheta^*+\bxi} (x_i) = 
J_{\btheta^*}^1(x_i) (\bDelta^1 x_i + \bdelta^1),\\
&\tilde a^l_{\btheta^*+\bxi} (x_i) = 
J_{\btheta^*}^l (x_i)
\left (
\bDelta^l a^{l-1}_{\btheta^*}(x_i) + \bdelta^l + (\bW_*^l+\bDelta^l)\tilde a^{l-1}_{\btheta^*+\bxi} (x_i)
\right).
\end{align*}
Similarly, let $\tilde f_{\btheta^*+\bxi}(\cdot)$ denote the amount of perturbation in the network output. It can be checked that
\begin{align*}
\tilde f_{\btheta^*+\bxi}(x_i) 
\defeq f_{\btheta^*+\bxi}(x_i) - f_{\btheta^*}(x_i) =
\bDelta^L a_{\btheta^*}^{L-1}(x_i) + \bdelta^L 
+(\bW_*^L + \bDelta^L) \tilde a_{\btheta^*+\bxi}^{L-1} (x_i).
\end{align*}
One can see that $\tilde a^1_{\btheta^*+\bxi} (x_i)$ only contains perturbation terms that are first-order in $\bxi$: $J_{\btheta^*}^1(x_i)(\bDelta^1 x_i + \bdelta^1)$.
However, the order of perturbation accumulates over layers. For example, 
\begin{align*}
\tilde a^2_{\btheta^*+\bxi} (x_i) 
&= 
J_{\btheta^*}^2 (x_i)
\left (
\bDelta^2 a^{1}_{\btheta^*}(x_i) + \bdelta^2 + (\bW_*^2+\bDelta^2)J_{\btheta^*}^1 (x_i) (\bDelta^1 x_i + \bdelta^1)
\right)\\
&=
\underbrace{
J_{\btheta^*}^2 (x_i)
\left (
\bDelta^2 a^{1}_{\btheta^*}(x_i) + \bdelta^2 + \bW_*^2 J_{\btheta^*}^1 (x_i)(\bDelta^1 x_i + \bdelta^1)
\right)
}_{\text{first-order perturbation}}
+
\underbrace{
J_{\btheta^*}^2 (x_i)
\bDelta^2 J_{\btheta^*}^1 (x_i) (\bDelta^1 x_i + \bdelta^1)
}_{\text{second-order perturbation}},
\end{align*}
so $\tilde a^2_{\btheta^*+\bxi} (x_i)$ contains 1st--2nd order perturbations. Similarly,
$\tilde a^l_{\btheta^*+\bxi} (x_i)$ has terms that are 1st--$l$-th order in $\bxi$, 
and $\tilde f_{\btheta^*+\bxi}(\cdot)$ perturbation terms from 1st order to $L$-th order.

Using the definition of $D_{\btheta}^l(x_i)$ from Eq~\eqref{eqn:Dmatrix}, the collection of first order perturbation terms in $\tilde f_{\btheta^*+\bxi}(\cdot)$ can be written as
\begin{align*}
\tilde f^{\textup{1}}_{\btheta^*+\bxi}(x_i) &\defeq 
\bDelta^L a_{\btheta^*}^{L-1}(x_i) + \bdelta^L + 
\bW_*^L J_{\btheta^*}^{L-1}(x_i)(\bDelta^{L-1} a_{\btheta^*}^{L-2}(x_i) + \bdelta^{L-1}) + \cdots\\
&= \sum_{l=1}^L 
D_{\btheta^*}^l (x_i) (\bDelta^l a_{\btheta^*}^{l-1} + \bdelta^l)
\stackrel{\text{(a)}}{=} \nabla_{\btheta} f_{\btheta^*} (x_i)^T \bxi 
\stackrel{\text{(b)}}{=} \nu_i^T \bxi_{\parallel}
\end{align*}
where (a) is an application of Taylor expansion of $f_{\btheta}(x_i)$ at $\btheta^*$, which can also be directly checked from explicit forms of $\bxi$~\eqref{eqn:thetaxivec} and $\nabla_\btheta f_{\btheta^*}(x_i)$~\eqref{eqn:gradfxi}.
Equality (b) comes from the definition of $\bxi_{\perp}$ that $\bxi_{\perp} \perp \nu_i$. We also define the collection of higher order perturbation terms:
\begin{equation*}
\tilde f^{\textup{2+}}_{\btheta^*+\bxi}(x_i) \defeq \tilde f_{\btheta^*+\bxi}(x_i) - \tilde f^{\textup{1}}_{\btheta^*+\bxi}(x_i).
\end{equation*}

Now, from the definition of memorizing global minima, $\ell'_i (\btheta^*) = 0$ for all $i \in [N]$. Since $\ell_i$ is three times differentiable, Taylor expansion of $\ell_i(\cdot)$ at $\btheta^*$ gives
\begin{equation*}
\ell_i(\btheta^*+\bxi) - \ell_i(\btheta^*) = \frac{1}{2} \ell''_i(\btheta^*) (\tilde f_{\btheta^*+\bxi}(x_i))^2 + \frac{1}{6} \alpha_i (\tilde f_{\btheta^*+\bxi}(x_i))^3,
\end{equation*}
where $\alpha_i = \ell'''(f_{\btheta^*}(x_i)+\beta_i \tilde f_{\btheta^*+\bxi} (x_i);y_i)$ for some $\beta_i \in [0,1]$.
For small enough $\rho_s$, $\tilde f_{\btheta^*+\bxi}(x_i)$ is small enough and bounded, so there exists a constant $C_1$ such that
\begin{equation*}
\ell_i(\btheta^*+\bxi) - \ell_i(\btheta^*) \leq C_1 (\tilde f_{\btheta^*+\bxi}(x_i))^2
\end{equation*}
for all $i \in [N]$.
There also are constants $C_2 \defeq \max_{i \in [N]} \norms{\nu_i}$ and $C_3$ such that 
\begin{align*}
|\tilde f^{\textup{1}}_{\btheta^*+\bxi}(x_i)| \leq C_2 \norms{\bxi_{\parallel}}, ~~\text{and}~~
|\tilde f^{\textup{2+}}_{\btheta^*+\bxi}(x_i)| \leq C_3 \norms{\bxi}^2
\end{align*}
for all $i \in [N]$, therefore 
\begin{equation*}
\ell(f_{\btheta^*+\bxi}(x_i);y_i) - \ell(f_{\btheta^*}(x_i);y_i) \leq C_1 (C_2 \norms{\bxi_{\parallel}} +C_3 \norms{\bxi}^2)^2
\end{equation*}
holds for all $i \in[N]$, as desired.

Now, consider the Taylor expansion of $\ell'_i$ at $f_{\btheta^*}(x_i)$. Because $\ell'_i$ is twice differentiable and $\ell'_i(\btheta^*) = 0$,
\begin{align}
\ell'_i(\btheta^*+\bxi)
=&
\ell''(\btheta^*) \tilde f_{\btheta^*+\bxi} (x_i) +
\frac{1}{2} \hat \alpha_i (\tilde f_{\btheta^*+\bxi} (x_i))^2 \nonumber \\
=&
\ell''(f_{\btheta^*}(x_i);y_i) \tilde f^{\textup{1}}_{\btheta^*+\bxi} (x_i) +
\underbrace{
	\ell''(f_{\btheta^*}(x_i);y_i) \tilde f^{\textup{2+}}_{\btheta^*+\bxi} (x_i) +
	\frac{1}{2} \hat \alpha_i (\tilde f_{\btheta^*+\bxi} (x_i))^2
}_{\eqdef R_i(\bxi)} \nonumber \\
=&
\ell''(f_{\btheta^*}(x_i);y_i) \nu_i^T \bxi_{\parallel} + R_i(\bxi), \label{eqn:taylorellp}
\end{align}
where $\hat \alpha_i = \half \ell'''(f_{\btheta^*}(x_i)+\hat \beta_i \tilde f_{\btheta^*+\bxi} (x_i);y_i)$ for some $\hat \beta_i \in [0,1]$.
The remainder term $R_i(\bxi)$ contains all the perturbation terms that are 2nd-order or higher, so there is a constant $C_4$ such that
\begin{equation*}
| R_i (\bxi) | \leq C_4 \norms{\bxi}^2
\end{equation*}
holds for all $i \in [N]$.

In a similar way, we can see from Eq~\eqref{eqn:gradfxi} that we can express $\nabla_{\btheta} f_{\btheta^*+\bxi} (x_i)$ as the sum of $\nu_i \defeq \nabla_{\btheta} f_{\btheta^*} (x_i)$ plus the perturbation $\mu_i(\bxi)$:
\begin{equation*}
\nabla_{\btheta} f_{\btheta^*+\bxi} (x_i) = \nu_i + \mu_i (\bxi),
\end{equation*}
where $\mu_i(\bxi)$ contains all the perturbation terms that are 1st-order or higher. So, there exists a constant $C_5$ such that
\begin{equation*}
\norms{ \mu_i (\bxi) } \leq C_5 \norms{\bxi}
\end{equation*}
holds for all $i \in [N]$.

%

\end{document}